\PassOptionsToPackage{table}{xcolor}
\documentclass[sigconf,nonacm]{acmart}
\usepackage{algorithm}
\usepackage{algorithmic}
\usepackage{amsmath}
\usepackage{array}
\usepackage{tabularx}
\usepackage{xurl}
\usepackage{multirow}
\usepackage{placeins}

\setkeys{acmart.cls}{balance=false}
\setcopyright{none}
\acmDOI{}
\acmISBN{}
\acmConference[WWW '27]{The ACM Web Conference 2027}{May 10--14, 2027}{Dublin, Ireland}
\acmYear{2027}
\copyrightyear{2027}

\definecolor{PaperText}{HTML}{20252B}
\definecolor{MutedText}{HTML}{4B5563}
\definecolor{SeedBlue}{HTML}{1766A3}
\definecolor{SeedBrightBlue}{HTML}{267BB4}
\definecolor{SeedCyan}{HTML}{007F7B}
\definecolor{SeedMint}{HTML}{007F7B}
\definecolor{BaselineGray}{HTML}{74777D}
\definecolor{StrokeBlue}{HTML}{1766A3}
\definecolor{StrokeTeal}{HTML}{007F7B}
\definecolor{BoardPurple}{HTML}{71539C}
\definecolor{BoardTint}{HTML}{F3EFF8}
\definecolor{TealTint}{HTML}{EDF7F5}
\definecolor{NeutralTint}{HTML}{F7F5F2}
\definecolor{BestTint}{HTML}{F0F1EE}
\definecolor{StrokeGray}{HTML}{4B5563}
\definecolor{CoralDark}{HTML}{C84135}
\definecolor{CoralTint}{HTML}{FFF1EE}
\definecolor{SeedTint}{HTML}{EAF3FB}
\definecolor{SeedTintStrong}{HTML}{D9E9F7}
\definecolor{HeaderGray}{HTML}{F0F0EE}
\definecolor{StripeGray}{HTML}{F7F6F3}
\definecolor{BorderGray}{HTML}{A8ADB2}
\definecolor{SoftGreen}{HTML}{F4F9F5}
\definecolor{SoftYellow}{HTML}{F4F8FB}
\definecolor{SoftRed}{HTML}{FBF4F2}
\definecolor{FakeRed}{HTML}{BA3641}
\definecolor{FakeTint}{HTML}{FBF4F2}
\definecolor{IconGreen}{HTML}{138457}
\definecolor{IconYellow}{HTML}{626B75}
\definecolor{IconRed}{HTML}{BA3641}

\newcommand{\bestscore}[1]{\textbf{#1}}
\newcommand{\posdelta}[1]{\textcolor{IconGreen}{#1}}
\newcommand{\negdelta}[1]{\textcolor{FakeRed}{#1}}
\newcommand{\zerodelta}[1]{\textcolor{MutedText}{#1}}

\def\signeddeltaaux#1#2\signeddeltaend{\ifx#1+\posdelta{#1#2}\else\ifx#1-\negdelta{#1#2}\else#1#2\fi\fi
}
\newcommand{\bestcell}[1]{\bestscore{#1}}

\newcommand{\selectortag}[3]{\textcolor{#1}{#2}\,\textcolor{PaperText}{#3}}
\newcommand{\textselector}{\selectortag{SeedBlue}{\ding{45}}{Text-indexed}}
\newcommand{\vlmselector}{\selectortag{BoardPurple}{\ding{72}}{VLM-board}}
\newcommand{\ocrselector}{\selectortag{StrokeTeal}{\ding{43}}{OCR+Text board}}

\newcommand{\evidencecards}{\textcolor{CoralDark}{\ding{111}}}
\newcommand{\evidenceboard}{\textcolor{StrokeTeal}{\ding{115}}}

\newcommand{\evabsent}{\textcolor{MutedText}{\bfseries --}}
\newcommand{\TopK}{\operatorname{TopK}}
\newcommand{\gridsetup}{\color{PaperText}\arrayrulecolor{BorderGray}\setlength{\arrayrulewidth}{0.4pt}\renewcommand{\arraystretch}{1.14}}
\newcommand{\gridreset}{\arrayrulecolor{black}\setlength{\arrayrulewidth}{0.4pt}}
\newcolumntype{Y}[1]{>{\raggedright\arraybackslash}m{#1}}

\newsavebox{\promptboxcontent}
\newsavebox{\findingboxcontent}
\newenvironment{findingbox}[1][Finding]{\begin{trivlist}\item\relax\begingroup\setlength{\fboxsep}{4pt}\begin{lrbox}{\findingboxcontent}\begin{minipage}{\dimexpr\linewidth-2\fboxsep-2\fboxrule\relax}\textcolor{CoralDark}{\textbf{#1.}}\enspace
}{\end{minipage}\end{lrbox}\noindent\fcolorbox{BorderGray}{NeutralTint}{\usebox{\findingboxcontent}}\endgroup\end{trivlist}
}

\usepackage[skins]{tcolorbox}
\usepackage{fontawesome5}
\definecolor{ArxivAbstractLavender}{HTML}{ECE7F5}
\acmConference{}{}{}
\acmBooktitle{}
\acmYear{2026}
\copyrightyear{2026}
\newcommand{\SkillLensDisplayTitle}{SkillLens: Visual Skill Cards for Retrieval-Augmented\\GUI Action Prediction and On-Policy Distillation}
\makeatletter
\def\@ACM@checkaffil{}
\def\@mktitle{\setbox\mktitle@bx=\vbox{\hsize=\textwidth
    \noindent\hbox to\textwidth{\raisebox{-0.5\height}{\includegraphics[height=0.40in]{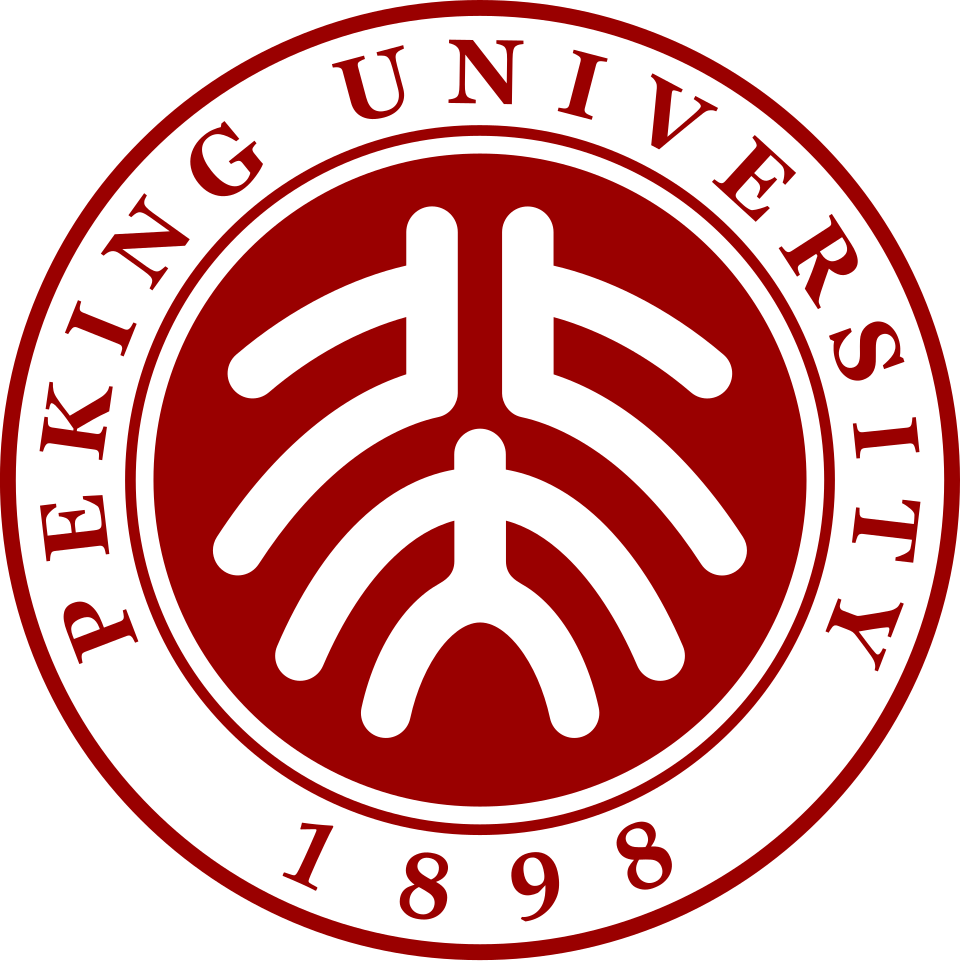}}\hspace{12pt}\raisebox{-0.5\height}{\includegraphics[height=0.40in]{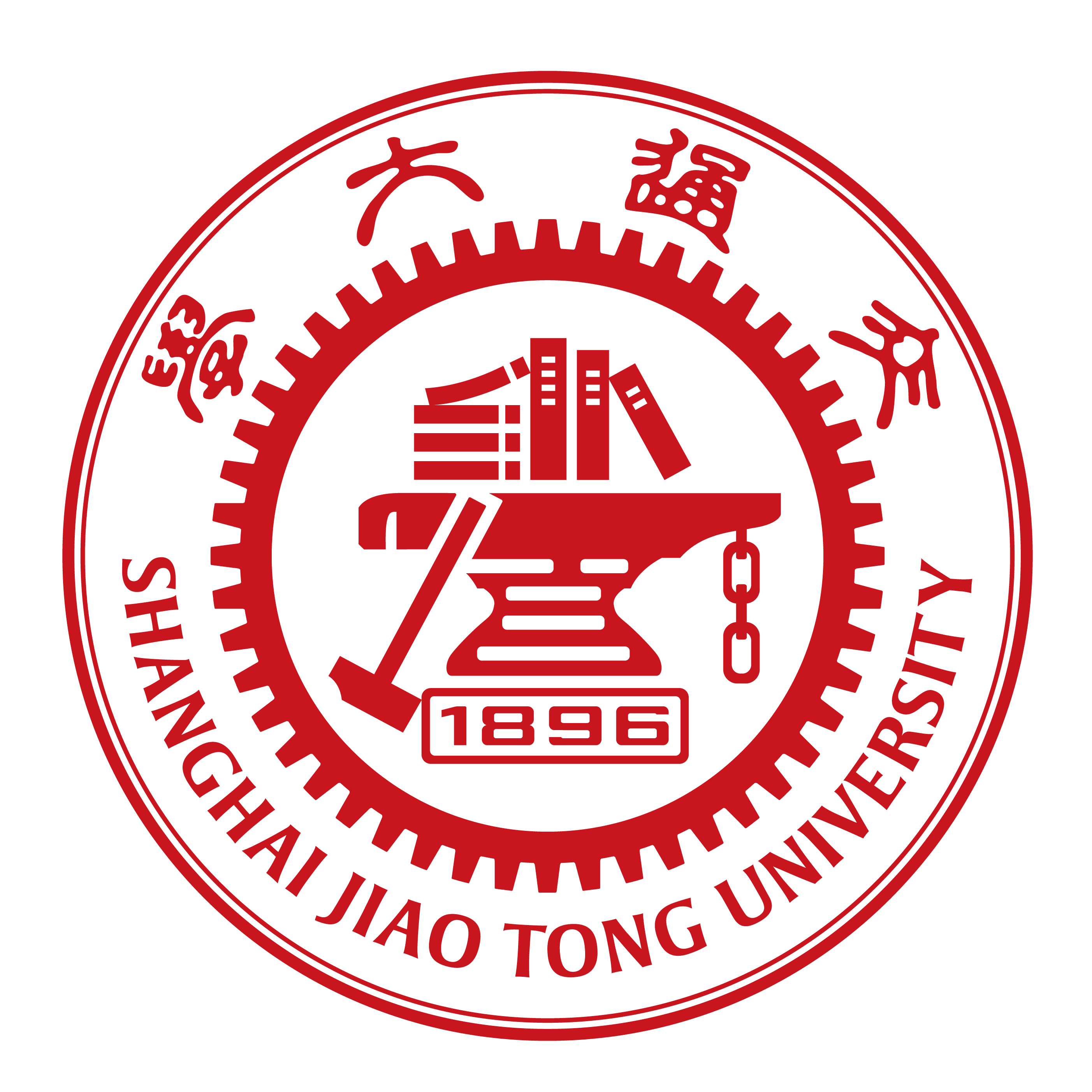}}\hspace{12pt}\raisebox{-0.5\height}{\includegraphics[width=1.55in]{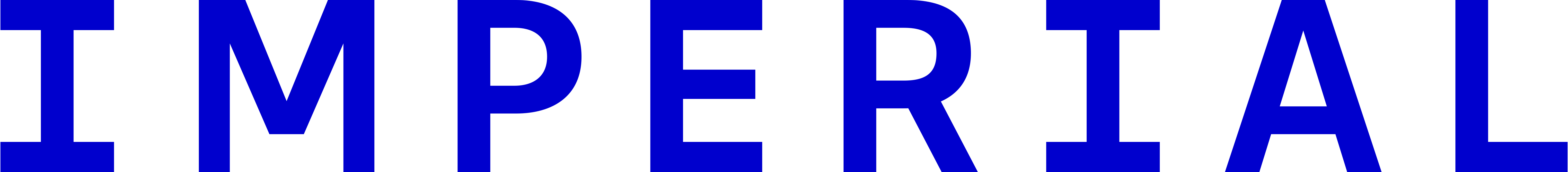}}\hspace{12pt}\raisebox{-0.5\height}{\includegraphics[height=0.40in]{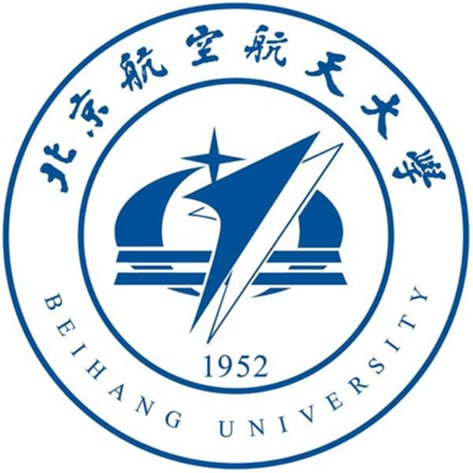}}\hfill
      \raisebox{-0.5\height}{\includegraphics[height=0.42in]{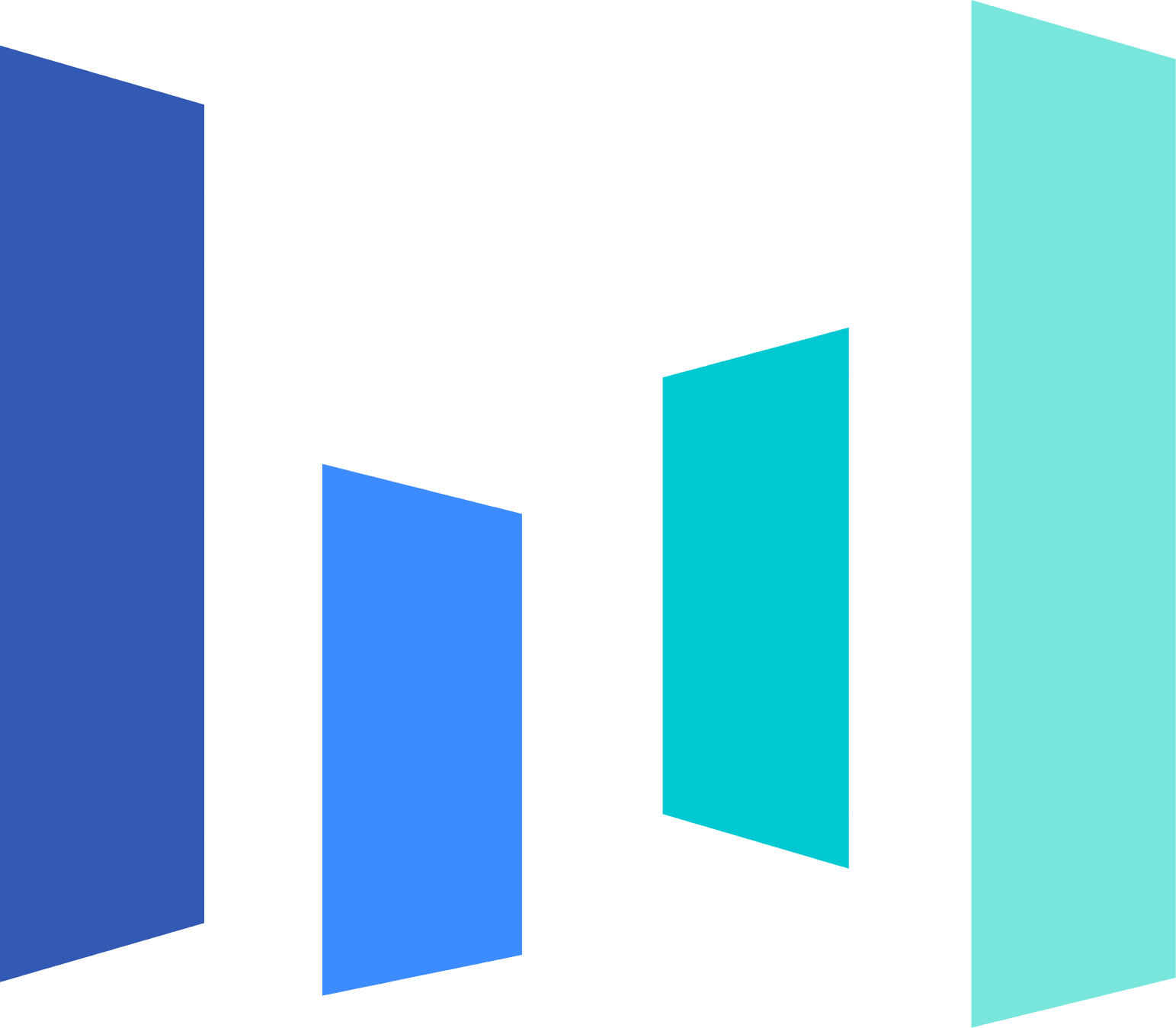}}}\par\vspace{10pt}\hrule height 0.65pt
    \vspace{8pt}{\centering\fontsize{17.5}{20.5}\selectfont\bfseries\SkillLensDisplayTitle\par}\vspace{8pt}\hrule height 0.65pt
  }}
\newbox\SkillLensTitleOnly
\let\SkillLensOriginalAuthors\@mkauthors
\def\@mkauthors{\setbox\SkillLensTitleOnly=\copy\mktitle@bx
  \SkillLensOriginalAuthors
  \global\setbox\mktitle@bx=\vbox{\hsize=\textwidth\unvbox\SkillLensTitleOnly
    \vspace{10pt}{\centering\fontsize{11.3}{14}\selectfont\bfseries
      Zhou Liu\textsuperscript{1,2,\ensuremath{\dagger}}\quad
      Chenxu Zhang\textsuperscript{4,\ensuremath{\dagger}}\quad
      Ligang Huang\textsuperscript{1}\quad
      Zeli Su\textsuperscript{1}\quad
      Zewei Pan\textsuperscript{3}\par
      \vspace{4pt}Changhang Tian\textsuperscript{5}\quad
      Zhaoyang Han\textsuperscript{1}\quad
      Xing Chen\textsuperscript{2}\quad
      Yuanfeng Song\textsuperscript{2}\quad
      Wentao Zhang\textsuperscript{1,*}\par}\vspace{7pt}{\centering\fontsize{10.5}{12.5}\selectfont
      \mbox{\textsuperscript{1}Peking University}\qquad
      \mbox{\textsuperscript{2}ByteDance}\qquad
      \mbox{\textsuperscript{3}Shanghai Jiao Tong University}\par
      \vspace{4pt}\mbox{\textsuperscript{4}Imperial College London}\qquad
      \mbox{\textsuperscript{5}Beihang University}\par}\vspace{7pt}{\centering\fontsize{8.5}{10}\selectfont
      \textsuperscript{\ensuremath{\dagger}}Equal contribution.\qquad
      \textsuperscript{*}Corresponding author.\par}\vspace{10pt}}}
\def\@printtopmatter{\ifx\@startPage\@empty\gdef\@startPage{1}\else\setcounter{page}{\@startPage}\fi
  \thispagestyle{plain}\noindent\box\mktitle@bx\par
}
\def\@mkabstract{\begingroup
  \let\@vspace\@vspace@orig\let\@vspacer\@vspacer@orig
  \begin{tcolorbox}[enhanced,colback=ArxivAbstractLavender,colframe=ArxivAbstractLavender,
    boxrule=0.6pt,arc=5pt,left=10pt,right=10pt,top=7pt,bottom=8pt,
    before skip=0pt,after skip=8pt,fontupper=\fontsize{9.5}{11.2}\selectfont]
    {\centering\fontsize{11}{13}\selectfont\bfseries Abstract\par}\vspace{4pt}\noindent\ignorespaces\@abstract\par
    \vspace{6pt}\noindent\faTags\hspace{0.45em}\textbf{Keywords:} \@keywords\par
  \end{tcolorbox}\endgroup
}
\patchcmd{\@printendtopmatter}{\par\bigskip}{\par}{}{}
\patchcmd{\maketitle}{\ifx\@keywords\@empty}{\iftrue}{}{\PackageError{skilllens-arxiv}{Could not suppress duplicate keywords}{}}
\makeatother

\title[SkillLens]{SkillLens: Visual Skill Cards for Retrieval-Augmented GUI Action Prediction and On-Policy Distillation}
\author{Zhou Liu}
\affiliation{\institution{Peking University}}
\affiliation{\institution{ByteDance}}
\author{Chenxu Zhang}
\affiliation{\institution{Imperial College London}}
\author{Ligang Huang}
\affiliation{\institution{Peking University}}
\author{Zeli Su}
\affiliation{\institution{Peking University}}
\author{Zewei Pan}
\affiliation{\institution{Shanghai Jiao Tong University}}
\author{Changhang Tian}
\affiliation{\institution{Beihang University}}
\author{Zhaoyang Han}
\affiliation{\institution{Peking University}}
\author{Xing Chen}
\affiliation{\institution{ByteDance}}
\author{Yuanfeng Song}
\affiliation{\institution{ByteDance}}
\author{Wentao Zhang}
\affiliation{\institution{Peking University}}
\renewcommand{\shortauthors}{Liu, Zhang, et al.}

\begin{document}

\begin{abstract}
Graphical user interface (GUI) agents can reuse past interaction experience. We study how illustrated procedures support action prediction and students deployed without retrieval. We introduce SkillLens, which represents procedures, applicability cues, and interface views in a shared Visual Skill Card (VSC) schema. Retrieved cards provide a frozen executor with reference evidence alongside the current screen. CardDistill uses this schema to construct privileged teacher context from labeled training steps for on-policy distillation; the student learns from native inputs and requires no retrieval at deployment. With matched training data and update counts, students trained with full-card teachers improve step success rate by an average of 5.04 percentage points over those trained with compact-text teachers across three training seeds on 119 tasks from ten websites absent from training. An independent runtime control uses training-source memory and pre-action retrieval queries: full cards achieve 12/64 correct actions (18.75\%) versus 4/64 (6.25\%) for text-only cards. These results establish the value of the complete VSC teacher-context package for retrieval-free students and provide evidence for pre-action retrieval of illustrated procedures.
\end{abstract}

\ccsdesc[500]{Information systems~Web applications}
\ccsdesc[300]{Computing methodologies~Intelligent agents}
\ccsdesc[300]{Information systems~Information retrieval}
\keywords{Web agents, GUI action prediction, visual skill cards, retrieval, on-policy distillation}

\onecolumn
\newgeometry{top=0.45in,bottom=0.45in,left=0.75in,right=0.75in,
  headheight=0pt,headsep=0pt,footskip=0.27in}
\maketitle
\pagestyle{plain}
\begingroup
\makeatletter
\let\@vspace\@vspace@orig\let\@vspacer\@vspacer@orig
\makeatother
\par\nointerlineskip
\noindent\begin{minipage}{\textwidth}
\captionsetup{type=figure,skip=6pt}
\centering
\includegraphics[width=\textwidth,trim=0 11bp 0 11bp,clip]{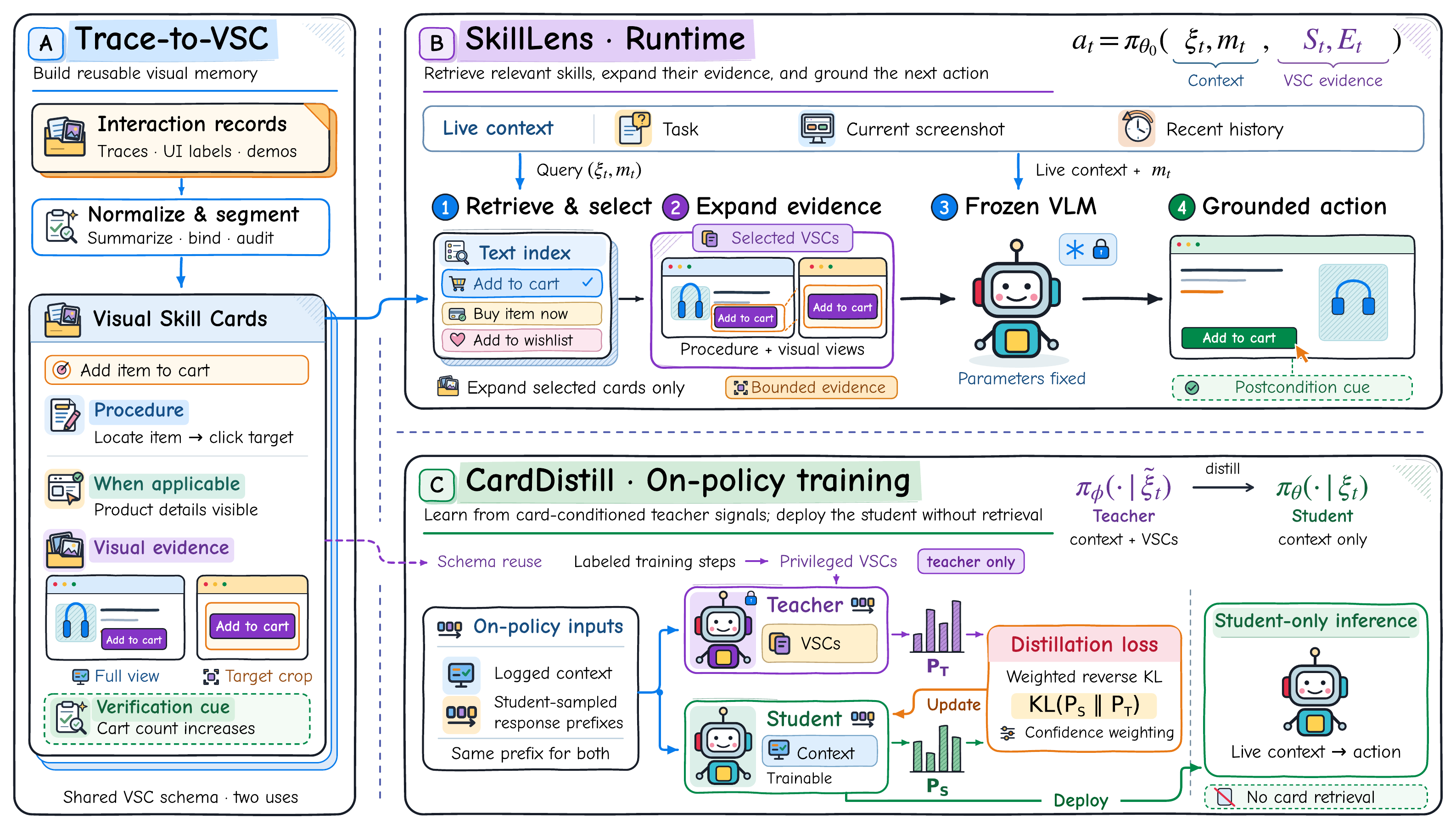}
\caption{SkillLens overview with Visual Skill Cards (VSCs). (A) Trace-to-VSC binds procedures, state cues, and visual evidence. (B) Retrieval and bounded expansion supply selected cards alongside the live screen to a frozen executor. (C) CardDistill trains on student-generated prefixes using privileged teacher cards; the deployed student needs no card retrieval.}
\Description{Three panels show visual skill card construction, retrieval and evidence expansion for a frozen executor, and on-policy teacher-student distillation. Only the teacher receives training cards; the student is deployed without card retrieval.}
\label{fig:system-overview}
\end{minipage}\par
\endgroup
\restoregeometry
\twocolumn
\setlength{\textfloatsep}{14pt}
\setlength{\dbltextfloatsep}{14pt}

\begingroup
\RenewEnviron{figure*}[1][]{\Description{The system overview and its description are presented on the opening page.}}
\section{Introduction}
Computer-using agents map requests to actions on graphical user interfaces (GUIs) using vision-language models (VLMs) to interpret screenshots~\citep{openai2024gpt4o,comanici2025gemini25,bai2024qwen2vl,bai2025qwen25vl,bai2025qwen3vl} and procedural knowledge to identify operations, applicable states, and signs of progress. Across layouts, agents must ground each action on the live interface (\mbox{Figure~\ref{fig:intro-case}}).

\begin{figure}[!htbp]
\centering
\includegraphics[width=\columnwidth]{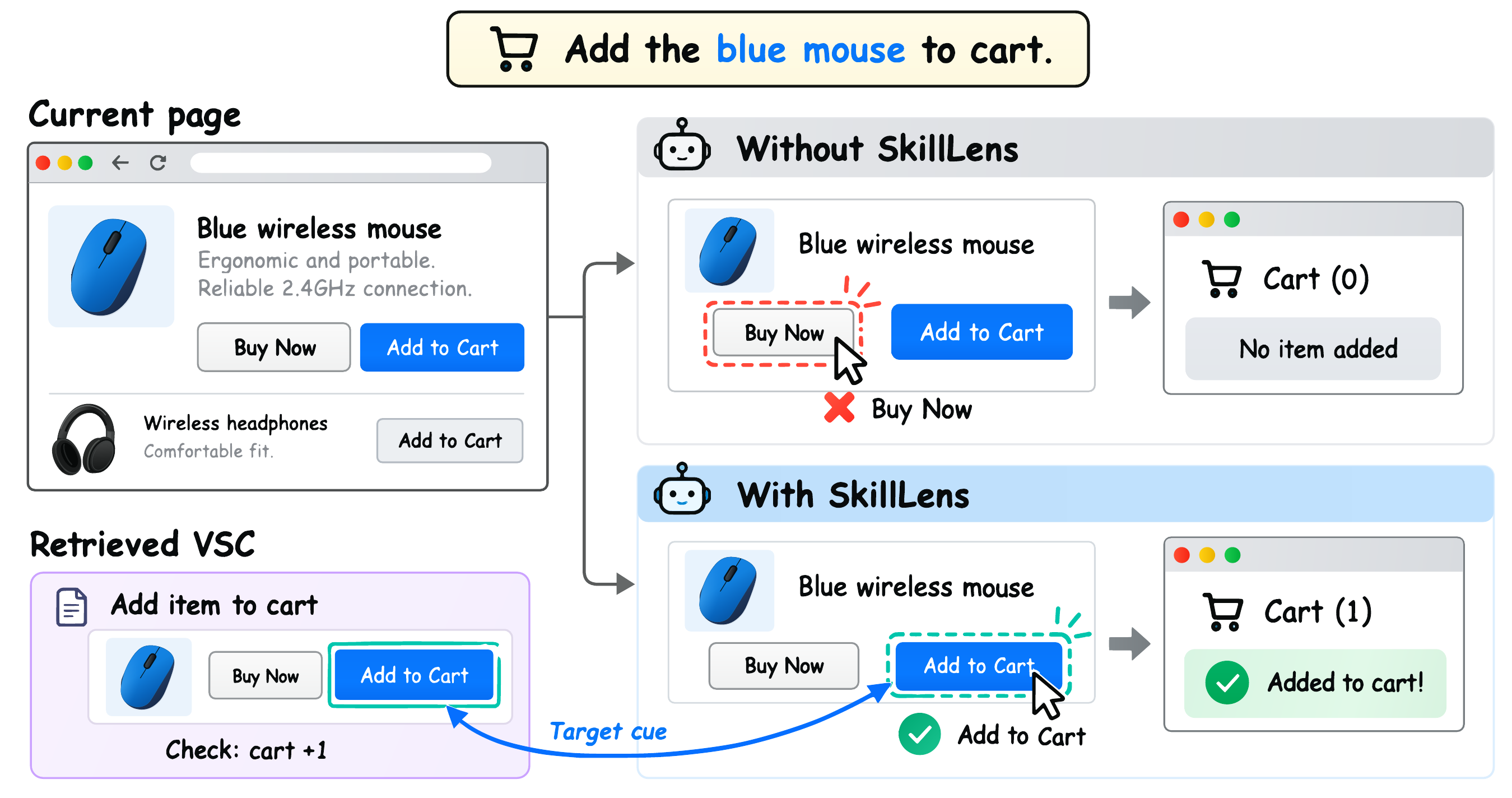}
\caption{Illustrative shopping example. A retrieved Visual Skill Card (VSC) links a target cue to the add-to-cart procedure, distinguishing it from the adjacent Buy Now control.}
\Description{A schematic shopping page shows a blue mouse and adjacent Buy Now and Add to Cart buttons. The no-skill path selects Buy Now and leaves the cart empty. The SkillLens path uses a retrieved visual skill card to select Add to Cart, with an expected cart count increase from zero to one.}
\label{fig:intro-case}
\end{figure}

Web agents encounter recurring workflows across websites: finding a product, applying a filter, or completing a form. Reusing these workflows requires both procedural knowledge and action grounding on pages whose controls and layouts change. Reusable skills store procedures~\citep{chen2026cuaskill,xie2025mirage,hu2026toolcua}; MMSkills and VISUALSKILL attach state information and visual evidence~\citep{zhang2026mmskills,jiang2026visualskill}. We study a memory interface that connects a retrieved procedure to its illustrated page context and supplies both to the agent making the current action decision. This connects reusable web experience to page-level grounding. We also ask whether learning from this interface improves a student's actions on unseen websites when deployed without skill retrieval.

We introduce \textbf{SkillLens} to study how visual skills support action prediction at inference time and through student training. Its \textbf{Visual Skill Cards (VSCs)} pair procedures with interface views and cues for applicability and verification. Trace-to-VSC constructs cards from recorded web actions and desktop grounding examples~\citep{deng2023mind2web,lu2024weblinx,xie2025osworldg}. At inference time, a text index selects relevant cards, whose procedures and visual evidence accompany the live screen. A frozen executor uses this context to predict the current action.

\begin{figure*}[t]
\centering
\includegraphics[width=.94\textwidth,trim=0 11bp 0 11bp,clip]{figures/SkillLens_Overview_Equations.pdf}
\caption{SkillLens overview with Visual Skill Cards (VSCs). (A) Trace-to-VSC binds procedures, state cues, and visual evidence. (B) Retrieval and bounded expansion supply selected cards alongside the live screen to a frozen executor. (C) CardDistill trains on student-generated prefixes using privileged teacher cards; the deployed student needs no card retrieval.}
\Description{Three panels show visual skill card construction, retrieval and evidence expansion for a frozen executor, and on-policy teacher-student distillation. Only the teacher receives training cards; the student is deployed without card retrieval.}
\label{fig:system-overview}
\end{figure*}

To test whether card evidence can also improve a student that acts without retrieval, we further develop \textbf{CardDistill}. It uses the same VSC schema to construct privileged teacher context from labeled training steps. During on-policy distillation, the student generates actions from task context and the live screen, and the teacher uses the additional card evidence to score tokens along these responses. The student learns from these scores and acts without cards.

We evaluate both uses of VSCs in offline studies on Multimodal-Mind2Web~\citep{deng2023mind2web,zheng2024seeact}, WebLINX-BrowserGym (WebLINX-BG)~\citep{lu2024weblinx,dechezelles2025browsergym}, and OSWorld-G~\citep{xie2025osworldg}. Runtime evidence spans benchmark-specific reference protocols and an independent training-source control. Full cards lead both single-modality variants in the Web reference comparisons and yield 12 strict successes versus four for text-only cards in the independent 64-task control. In controlled training, full-card teachers gain 5.04 points in step success rate (Step SR) over compact text across three seeds on 119 tasks from ten training-disjoint websites.

{In summary, our contributions are threefold:}

\begingroup
\settowidth{\leftmargini}{\labelitemi}
\addtolength{\leftmargini}{\labelsep}
\begin{itemize}
\item We propose SkillLens, a VSC memory interface that connects reusable procedures and visual evidence to action prediction by frozen executors.
\item We develop CardDistill to use the same VSC schema for privileged teacher context and on-policy training of students deployed without card retrieval.
\item We demonstrate retrieval-free student gains under controlled training across three seeds and on training-disjoint websites, and provide an independent pre-action runtime comparison using training-source memory.
\end{itemize}
\endgroup

\endgroup
\section{Related Work}
\label{sec:related-work}

\noindent\textbf{Procedural memory for visual agents.}
Reusable skills encode procedures and tool-use rules~\citep{chen2026cuaskill,xie2025mirage,hu2026toolcua}. MMSkills packages procedures, state cards, and keyframes for branch-based consultation~\citep{zhang2026mmskills}; VISUALSKILL builds illustrated guides for applications with on-demand loading~\citep{jiang2026visualskill}. Optical memory systems organize history as rendered segments and selectively loaded views~\citep{li2026ocrmemory,feng2026agentocr,shi2026memocr,liang2026vizomem,xie2026lensvlm,wei2025deepseek}. SkillLens binds procedures and page evidence at the action level, with text selection and bounded visual expansion. The same schema also organizes teacher supervision for parameter learning.

\noindent\textbf{Grounding and privileged visual supervision.}
GUI grounding models learn control localization~\citep{cheng2024seeclick,wu2024osatlas,gou2025uground,wu2025guiactor,xie2025osworldg}. GUI-SD and Vision-OPD transfer privileged visual information through on-policy distillation using target-focused masks and image crops~\citep{zhang2026guisd,yuan2026visionopd}; Imagine-OPD internalizes privileged zoomed evidence through on-policy self-distillation~\citep{cai2026imagineopd}. CardDistill organizes procedure, applicability, and target evidence in VSCs instantiated from labeled training steps. Student-generated actions receive card-conditioned teacher supervision, linking this representation to retrieval-free deployment. Our comparisons examine the complete teacher-context package across seeds and held-out websites.

\section{Method}
\label{sec:method}

Figure~\ref{fig:system-overview} shows two uses of the VSC schema: retrieved procedural evidence for frozen executors and privileged teacher context for student training. Trace-to-VSC builds the runtime library; CardDistill instantiates training cards from labeled steps. At step $t$, the native task context is $\xi_t=(x,o_t,h_t)$: the instruction, current screenshot or structured observation, and available interaction history. An action $a_t$ specifies an operation and its grounded target or arguments.

A web action identifies a candidate element and its operation or text value; a desktop grounding action identifies a screen point.

\subsection{Visual Skill Card Representation}
\label{sec:vsc-construction}
A Visual Skill Card is a tuple $s_i=(p_i,z_i,v_i,\kappa_i)$. Procedure text $p_i$ describes an operation or workflow; state cues $z_i$ specify applicability and observable progress; visual views $v_i$ show the interface and, when available, a focused target; auxiliary fields $\kappa_i$ store recovery cues or compact memory views. These fields connect what to do with the interface state in which the procedure is useful. For example, a shopping card can pair an ``add to cart'' procedure with the product-detail state, a full view and target crop, and a cart-count verification cue.

Each card stores a stable identifier, name, goal, ordered procedure, keywords, and source and modality metadata. A compact descriptor supports retrieval; the associated procedure and available images support action prediction. Shared identifiers bind both views, allowing the selector to search text before the loader reads image assets.

Multi-step traces can produce meta cards for task patterns, core cards for subgoals, and execution cards for short operations. The reported runtime experiments select execution-level cards.

Evidence availability follows the source record: grounded screenshots support target views, while longer traces can describe subgoals and state transitions. Target metadata, crops, and after-state images occupy distinct fields; the loader expands the available views listed in the library profile. Goal fields separate the procedure from its demonstrated layout.

\begin{figure}[!t]
\centering
\includegraphics[width=\columnwidth]{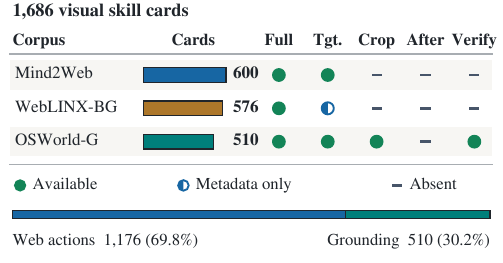}
\caption{Visual Skill Card (VSC) library. Filled/half-filled circles mark available evidence/metadata only. Full/Tgt.: full screenshot/target; dashes: absence.}
\Description{Counts of cards from three corpora, the available evidence views, and their web-action and desktop-grounding evaluation roles.}
\label{fig:vsc-corpus-profile}
\end{figure}

\subsection{Trace-to-VSC Construction}
\label{sec:trace-to-vsc}
The library contains 1,686 execution cards: 600 web-action cards from multimodal Mind2Web records~\citep{deng2023mind2web}, 576 dialogue-conditioned browser cards from WebLINX-BG~\citep{lu2024weblinx}, and 510 grounding cards from OSWorld-G~\citep{xie2025osworldg}. Mind2Web associates tasks and actions with page observations and target elements; WebLINX-BG adds dialogue and action history; OSWorld-G pairs an instruction and screenshot with target geometry. Figure~\ref{fig:vsc-corpus-profile} summarizes the library and available evidence.

Trace-to-VSC applies five steps: normalization, segmentation into reusable units, procedure summarization, evidence binding, and card validation. The structured-source adapters populate procedure and state fields from recorded instructions, actions, and annotations. A card corresponds to a source action or grounding example. Its source identifiers retain the link to that record; its exported text organizes the operation and supporting context for reuse.

Normalization associates each action with its screenshot and available page or target annotations. For Mind2Web, the adapter retains website and domain metadata, task text, recent actions, an HTML excerpt, and annotated target identifiers. WebLINX-BG cards retain the source demonstration and turn, dialogue, action history, page excerpt, and target action. Procedure fields describe the operation and its arguments, while the accompanying context describes the demonstrated interface state. The export stores readable text, a structured record, image captions, and asset checksums under a stable card identifier. Validation checks the record structure and the availability of linked assets.

Evidence binding connects these fields to image assets. For a source with a target box, a focused crop exposes the target's appearance while the full screenshot retains its surrounding layout. The exported procedure and image captions describe the role of each view, and the manifest binds their paths to the card identifier. For example, the OSWorld-G adapter exports a full screenshot, an annotated target view, and a crop, together with grounding steps and success checks. Figure~\ref{fig:vsc-corpus-profile} records which views each source provides, guiding the loader's selection of full-screen, target, and verification evidence. The grounding checks cover point placement inside the reference target and agreement among instruction, text, icon, and region cues.

\subsection{Retrieval and Evidence Loading}
\label{sec:runtime-retrieval}
Web adapters form queries from task and history context together with protocol-specific reference context $m_t$, whose inputs are defined in Section~\ref{sec:runtime-evaluation-settings}; $m_t$ is empty in the independent pre-action control. WebLINX-BG uses dialogue, action history, viewport information, and a page-text excerpt. Coarse retrieval matches queries to card descriptors. The reference ranking prioritizes site compatibility, operation match, and lexical score. OSWorld-G directly ranks cards using instruction and GUI-type tokens, with a domain-match bonus. Appendix~\ref{app:retrieval-details} details the reference implementation.

On the web path, three card budgets act at different stages. Coarse retrieval retains up to $K_c$ candidates from the library, reranking selects up to $K_s$ cards, and at most $K_e$ of these are considered for evidence expansion. An optional compatibility gate then filters this capped set. The Mind2Web gate accepts same-site cards and domain-matched cards with a compatible operation, using the operation information defined in Section~\ref{sec:runtime-evaluation-settings}. The loader resolves the remaining identifiers to stored procedures, image paths, roles, and captions. It attaches available reference screenshots and target views up to the image budget; the current screen is supplied separately.

More explicitly, let $r_i^{(0)}$ and $r_i^{(1)}$ be the coarse and reranking scores for card $s_i$, and let $\chi_i$ denote the optional compatibility gate. The web path forms
\begin{equation}
\begin{aligned}
\mathcal C_t&=\TopK_{K_c}(\mathcal L;r^{(0)}),\\
\widetilde S_t&=\TopK_{K_s}(\mathcal C_t;r^{(1)}),\\
S_t&=\{s_i\in\TopK_{K_e}(\widetilde S_t;r^{(1)}):\chi_i=1\}.
\end{aligned}
\label{eq:selection-stages}
\end{equation}
The gate defaults to one when disabled. Applying it after the expansion cap means that the final set may contain fewer than $K_e$ cards. In the shared text reranker, distinct query-token overlap is combined with $0.1$ times the incoming coarse score. Appendix~\ref{app:retrieval-details} gives the descriptor fields and separates this rule from adapter-specific coarse ranking.

Let $\rho_{\mathrm{sel}}$ denote this selector and $e$ the deterministic map from card identifiers to their original assets. The runtime interface is
\begin{equation}
\begin{aligned}
S_t&=\rho_{\mathrm{sel}}(\xi_t,m_t;\mathcal L), & E_t&=e(S_t),\\
a_t&=\pi_{\theta_0}(\xi_t,m_t,S_t,E_t), & |S_t|&\le K_e,
\end{aligned}
\label{eq:runtime-interface}
\end{equation}
with at most $K_v$ evidence images per decision and fixed executor parameters $\theta_0$. Both the selector and executor receive the reference context $m_t$ defined in Section~\ref{sec:runtime-evaluation-settings}.

We also evaluate two rendered-board selectors. VLM-board asks a VLM to choose a candidate row; OCR+Text board transcribes the rows and reranks their text. Both resolve the chosen row to a stable card identifier and then load the same original assets. This shared interface lets the experiments examine card selection separately from the executor's use of the retrieved evidence.

\subsection{Grounded Action Prediction}
\label{sec:grounded-action}
The executor prompt combines the instruction and history, current observation, expanded cards, reference context $m_t$ (Section~\ref{sec:runtime-evaluation-settings}), and the benchmark action schema. Procedure text supplies an operation-level cue, while the reference views show the interface context and target appearance that motivated it. The executor uses these cues to predict a target in the current observation. For example, an add-to-cart card describes a control's role and expected state change; the executor uses the current product page to choose the displayed control. Source annotations explain the reference view, and live-screen grounding determines the output target and coordinates.

Verification fields describe observable postconditions, such as an updated cart count, for checking progress in a later observation. The reported offline evaluations score the current action or target at a recorded state; verification cues describe checks on subsequent observations.

\begingroup\begin{table*}[!t]
\caption{Reference-protocol runtime results (\%; Section~\ref{sec:experimental-setup}); $\Delta$ in pp versus No-skill. Coral marks SkillLens; bold marks the higher score in each pair, including ties.}
\label{tab:main-web-results}
\centering
\normalsize
\setlength{\tabcolsep}{3.4pt}
\gridsetup
\renewcommand{\arraystretch}{1.0}
\color{black}
\begin{tabular}{|l|l|r|r|r|r|r|r|r|r|r|}
\hline
\rowcolor{HeaderGray}
\textbf{Model} & \textbf{Condition} &
\multicolumn{3}{c|}{\textbf{Multimodal-Mind2Web}} &
\multicolumn{3}{c|}{\textbf{WebLINX-BG}} &
\multicolumn{3}{c|}{\textbf{OSWorld-G}} \\
\cline{3-11}
\rowcolor{HeaderGray}
& &
\textbf{\shortstack[r]{Step\\SR $\uparrow$}} &
\textbf{\shortstack[r]{Action\\F1 $\uparrow$}} &
\textbf{\shortstack[r]{Element\\Acc. $\uparrow$}} &
\textbf{\shortstack[r]{Overall\\Score $\uparrow$}} &
\textbf{\shortstack[r]{Intent\\Match $\uparrow$}} &
\textbf{\shortstack[r]{Element\\IoU $\uparrow$}} &
\textbf{\shortstack[r]{Ground.\\Acc. $\uparrow$}} &
\textbf{\shortstack[r]{BBox\\Hit $\uparrow$}} &
\textbf{\shortstack[r]{Valid\\Output $\uparrow$}} \\
\hline
\multirow{3}{*}{\textbf{Qwen2.5-VL-3B}} & \evabsent~No-skill & 3.3 & 18.2 & 21.8 & 13.2 & 48.1 & 13.2 & 15.5 & 14.5 & 68.0 \\
& \cellcolor{CoralTint}\textcolor{CoralDark}{\ding{111}~\bfseries SkillLens} & \cellcolor{CoralTint}\bestscore{7.6} & \cellcolor{CoralTint}\bestscore{21.6} & \cellcolor{CoralTint}\bestscore{32.2} & \cellcolor{CoralTint}\bestscore{14.3} & \cellcolor{CoralTint}\bestscore{67.0} & \cellcolor{CoralTint}\bestscore{19.1} & \cellcolor{CoralTint}\bestscore{32.5} & \cellcolor{CoralTint}\bestscore{30.0} & \cellcolor{CoralTint}\bestscore{99.5} \\
& \textcolor{MutedText}{$\Delta$} & \posdelta{+4.3} & \posdelta{+3.4} & \posdelta{+10.4} & \posdelta{+1.1} & \posdelta{+18.9} & \posdelta{+5.9} & \posdelta{+17.0} & \posdelta{+15.5} & \posdelta{+31.5} \\
\hline
\multirow{3}{*}{\textbf{Qwen3-VL-2B}} & \evabsent~No-skill & 4.6 & 49.6 & 34.4 & 14.1 & 56.8 & 17.9 & 28.5 & 26.5 & 99.5 \\
& \cellcolor{CoralTint}\textcolor{CoralDark}{\ding{111}~\bfseries SkillLens} & \cellcolor{CoralTint}\bestscore{18.5} & \cellcolor{CoralTint}\bestscore{61.0} & \cellcolor{CoralTint}\bestscore{73.0} & \cellcolor{CoralTint}\bestscore{16.5} & \cellcolor{CoralTint}\bestscore{66.5} & \cellcolor{CoralTint}\bestscore{21.0} & \cellcolor{CoralTint}\bestscore{41.0} & \cellcolor{CoralTint}\bestscore{38.0} & \cellcolor{CoralTint}\bestscore{100.0} \\
& \textcolor{MutedText}{$\Delta$} & \posdelta{+13.9} & \posdelta{+11.4} & \posdelta{+38.6} & \posdelta{+2.4} & \posdelta{+9.7} & \posdelta{+3.1} & \posdelta{+12.5} & \posdelta{+11.5} & \posdelta{+0.5} \\
\hline
\multirow{3}{*}{\textbf{Gemini 3.1 Flash-Lite}} & \evabsent~No-skill & 57.8 & 75.9 & 70.6 & 7.8 & 49.5 & 11.3 & 15.5 & 13.0 & \bestscore{100.0} \\
& \cellcolor{CoralTint}\textcolor{CoralDark}{\ding{111}~\bfseries SkillLens} & \cellcolor{CoralTint}\bestscore{65.8} & \cellcolor{CoralTint}\bestscore{79.4} & \cellcolor{CoralTint}\bestscore{77.2} & \cellcolor{CoralTint}\bestscore{12.1} & \cellcolor{CoralTint}\bestscore{60.7} & \cellcolor{CoralTint}\bestscore{16.7} & \cellcolor{CoralTint}\bestscore{27.5} & \cellcolor{CoralTint}\bestscore{26.0} & \cellcolor{CoralTint}\bestscore{100.0} \\
& \textcolor{MutedText}{$\Delta$} & \posdelta{+8.0} & \posdelta{+3.5} & \posdelta{+6.6} & \posdelta{+4.3} & \posdelta{+11.2} & \posdelta{+5.4} & \posdelta{+12.0} & \posdelta{+13.0} & \zerodelta{+0.0} \\
\hline
\multirow{3}{*}{\textbf{Gemini 2.5 Flash}} & \evabsent~No-skill & 57.9 & 75.7 & 69.8 & 7.6 & 49.0 & \bestscore{11.0} & 17.5 & 16.0 & \bestscore{100.0} \\
& \cellcolor{CoralTint}\textcolor{CoralDark}{\ding{111}~\bfseries SkillLens} & \cellcolor{CoralTint}\bestscore{79.0} & \cellcolor{CoralTint}\bestscore{88.0} & \cellcolor{CoralTint}\bestscore{89.5} & \cellcolor{CoralTint}\bestscore{16.0} & \cellcolor{CoralTint}\bestscore{64.5} & \cellcolor{CoralTint}\bestscore{11.0} & \cellcolor{CoralTint}\bestscore{27.5} & \cellcolor{CoralTint}\bestscore{26.0} & \cellcolor{CoralTint}\bestscore{100.0} \\
& \textcolor{MutedText}{$\Delta$} & \posdelta{+21.1} & \posdelta{+12.3} & \posdelta{+19.7} & \posdelta{+8.4} & \posdelta{+15.5} & \zerodelta{+0.0} & \posdelta{+10.0} & \posdelta{+10.0} & \zerodelta{+0.0} \\
\hline
\multirow{3}{*}{\textbf{GPT-4o}} & \evabsent~No-skill & 65.7 & 92.3 & 77.5 & 13.5 & 59.7 & 19.9 & 1.5 & 1.5 & 86.5 \\
& \cellcolor{CoralTint}\textcolor{CoralDark}{\ding{111}~\bfseries SkillLens} & \cellcolor{CoralTint}\bestscore{82.4} & \cellcolor{CoralTint}\bestscore{94.4} & \cellcolor{CoralTint}\bestscore{87.1} & \cellcolor{CoralTint}\bestscore{15.3} & \cellcolor{CoralTint}\bestscore{66.0} & \cellcolor{CoralTint}\bestscore{22.1} & \cellcolor{CoralTint}\bestscore{3.5} & \cellcolor{CoralTint}\bestscore{3.5} & \cellcolor{CoralTint}\bestscore{89.5} \\
& \textcolor{MutedText}{$\Delta$} & \posdelta{+16.7} & \posdelta{+2.1} & \posdelta{+9.6} & \posdelta{+1.8} & \posdelta{+6.3} & \posdelta{+2.2} & \posdelta{+2.0} & \posdelta{+2.0} & \posdelta{+3.0} \\
\hline
\multirow{3}{*}{\textbf{GPT-5.4-mini}} & \evabsent~No-skill & 77.2 & 90.1 & 88.1 & 12.8 & 55.3 & 17.3 & 45.0 & 43.0 & \bestscore{100.0} \\
& \cellcolor{CoralTint}\textcolor{CoralDark}{\ding{111}~\bfseries SkillLens} & \cellcolor{CoralTint}\bestscore{89.1} & \cellcolor{CoralTint}\bestscore{96.4} & \cellcolor{CoralTint}\bestscore{92.0} & \cellcolor{CoralTint}\bestscore{15.7} & \cellcolor{CoralTint}\bestscore{67.0} & \cellcolor{CoralTint}\bestscore{19.0} & \cellcolor{CoralTint}\bestscore{66.5} & \cellcolor{CoralTint}\bestscore{65.0} & \cellcolor{CoralTint}\bestscore{100.0} \\
& \textcolor{MutedText}{$\Delta$} & \posdelta{+11.9} & \posdelta{+6.3} & \posdelta{+3.9} & \posdelta{+2.9} & \posdelta{+11.7} & \posdelta{+1.7} & \posdelta{+21.5} & \posdelta{+22.0} & \zerodelta{+0.0} \\
\hline
\end{tabular}
\gridreset
\end{table*}
\endgroup

\subsection{CardDistill with Privileged VSCs}
\label{sec:carddistill}
CardDistill uses on-policy distillation~\citep{zhang2026guisd,yuan2026visionopd} with card-conditioned teachers. The teacher observes $\tilde\xi_t=(\xi_t,S_t^{\mathrm{priv}},E_t^{\mathrm{priv}})$; the student receives $\xi_t$. In the controlled studies, cards are constructed from the labeled training step, providing privileged procedure and target context. Their reference screenshot is the current training screenshot; target-focused views and the rendered card expose additional label-derived evidence. This training path instantiates the VSC schema with current-example supervision, while runtime memory uses a library of prior experience. The student samples a response $\hat y_t\sim\pi_\theta(\cdot\mid\xi_t)$, and both policies score its prefixes $\hat y_{t,<j}$.

Both branches condition on identical generated prefixes, with aligned response tokens and masked prompt positions; gradients flow through the student forward pass.

Let $P^{(T)}_{\theta,t,j}$ and $P^{(T)}_{\phi,t,j}$ be the student and teacher token distributions at temperature $T$. Following GUI-SD, the objective weights reverse Kullback--Leibler (KL) divergence by teacher confidence and averages over $N$ valid response tokens:
\begin{equation}
\begin{aligned}
\mathcal L_{\mathrm{CD}}
&=\frac{1}{N}\sum_{t,j}w_{t,j}\,\mathrm{KL}\!\left(P^{(T)}_{\theta,t,j}\middle\|P^{(T)}_{\phi,t,j}\right),\\
w_{t,j}&=1-\frac{H(P^{(1)}_{\phi,t,j})}{\log|\mathcal V|},
\end{aligned}
\label{eq:carddistill-loss}
\end{equation}
where $\mathcal V$ is the output vocabulary. The normalized entropy gives a weight near one for a concentrated teacher distribution and near zero for a nearly uniform distribution, reducing uncertain token-level supervision. The reported runs use $T=1.0$.

Reverse KL penalizes student probability on tokens disfavored by the teacher; the entropy weight sets the strength of this alignment.

On-policy sampling here generates responses at logged training states. The screenshot and task remain fixed, while the student samples operation, target, and argument tokens. Card evidence enters the teacher context; the student retains its native input interface at training and evaluation.

In the matched studies, teacher scoring runs without gradients and the teacher weights remain fixed; exponential moving-average updates are disabled. Student adaptation uses low-rank adaptation (LoRA) with rank 8 and scaling factor 32 on language-model attention and feed-forward projections; the vision encoder and aligner are frozen. At deployment, the student predicts directly from native context; the teacher, card retrieval, and evidence expansion are absent. Appendix~\ref{app:submission-training} records the training configuration and pseudocode.

\section{Experiments}
\label{sec:experiments}

Our central evidence comes from controlled CardDistill training and student transfer to unseen websites (RQ4), complemented by an independent pre-action runtime control (RQ2). We also compare benchmark-specific reference configurations (RQ1), examine their card content (RQ2), and analyze selection diagnostics (RQ3).

\subsection{Experimental Setup}
\label{sec:experimental-setup}

\noindent\textbf{Benchmarks and executors.}
We evaluate offline action prediction on Multimodal-Mind2Web with candidate elements~\citep{deng2023mind2web,zheng2024seeact} and conversational WebLINX-BG turns~\citep{lu2024weblinx,dechezelles2025browsergym}, plus desktop target localization on OSWorld-G~\citep{xie2025osworldg}. Table~\ref{tab:main-web-results} covers six frozen backbones.

\noindent\textbf{Runtime evaluation settings.}
\label{sec:runtime-evaluation-settings}
We evaluate benchmark-specific reference configurations and an independent pre-action retrieval control. In the Mind2Web reference adapter, annotated current-action descriptions and operation labels inform card selection, ranking, and compatibility filtering. The memory-augmented executor prompt additionally includes the current-action description, which is absent from the No-skill prompt. The independent control retrieves from training-source memory using pre-action task and interface information. All conditions within each comparison share the evaluation cases, native task-and-screen inputs, and scoring implementation.

The WebLINX-BG retrieval-only diagnostic includes the labeled current intent in its query. The OSWorld-G coverage diagnostic recovers same-instance cards by case identifier.

\noindent\textbf{Memory conditions.}
Runtime reference studies use the 1,686-card library in Figure~\ref{fig:vsc-corpus-profile}. Within each Web comparison, Full, Text-only, Images-only, and No-skill share evaluation cases, model version, base task-and-screen prompt template, and scoring implementation. Additional reference context follows the runtime evaluation settings above; card text and images vary by memory condition. All conditions retain the native task context and live screen. The independent control uses training-source memory.

\noindent\textbf{Controlled CardDistill training.}
We fix Qwen3-VL-2B, 512 labeled Mind2Web training examples, adapter and optimizer settings, and 128 updates. Compact-text and Full-VSC teachers share the live screenshot and compact procedure fields; Full-VSC adds source-aligned visual evidence and a rendered card. These cards encode the current labeled training step as privileged supervision. Reference screenshots reproduce the training screen; target crops expose the annotated control. Richer text adds every rendered text line without extra images, and the no-rendered-card control removes the rendered card from Full-VSC. The controls match data and updates and compare teacher-context packages combining visual focus, layout, parameter binding, and context length. Full-VSC and compact text use seeds 42/43/44; richer text and no-rendered-card use seed 42. Students receive native live inputs without cards at evaluation.

\noindent\textbf{Student evaluation cohorts.}
The controlled \texttt{test\_task} cohort contains 200 steps from 120 tasks. The additional \texttt{test\_website} cohort contains 119 tasks, one step each, on ten websites absent from the 73 training sites. Conditions and checkpoints are fixed before cross-website inference. Across teacher-context conditions, students receive the same task context, live screenshot, and top-16 DeBERTa-ranked element candidates; all 30 cross-website cases whose annotated target is outside the shortlist remain in the cohort. We report the original Mind2Web and WebLINX-BG distillation cohorts separately (Appendix~\ref{app:carddistill-reliability}). External grounding comparisons convert native outputs to a shared center-in-target-box score on 200 examples per benchmark.

\noindent\textbf{Evaluation.}
Primary metrics are Mind2Web Step SR, WebLINX-BG Overall, and OSWorld-G Grounding Acc. (Appendix~\ref{app:submission-metrics}). Scores are percentages and differences are percentage points unless stated otherwise. Main-table gains use each executor's No-skill baseline. Controlled students share greedy decoding, parsing, and scoring. We report training-seed standard deviations separately from evaluation-sample confidence intervals. Cross-website comparisons use paired task bootstraps and a ten-website cluster bootstrap, with leave-one-website-out and pre-flagged near-similarity exclusions as sensitivity checks. Appendix~\ref{app:teacher-modality-control} details sampling and scoring.

\subsection{RQ1: How Do the Reference Configurations Compare?}
\label{sec:runtime-results}

\noindent\textbf{The reference configurations have higher primary point scores.}
Table~\ref{tab:main-web-results} reports 18 positive primary point differences for the complete reference configurations defined in Section~\ref{sec:runtime-evaluation-settings}. Differences span 4.3--21.1 points on Mind2Web, 1.1--8.4 on WebLINX-BG, and 2.0--21.5 on OSWorld-G. GPT-5.4-mini's reference configuration scores 89.1 versus 77.2 Step SR, 15.7 versus 12.8 Overall, and 66.5 versus 45.0 Grounding Acc. Qwen2.5-VL-3B's reference configuration scores 32.5 versus 15.5 in grounding (+17.0 points).

\begingroup\begin{table*}[t]
\caption{Card content comparisons (\%), retaining the live task and screen. $\Delta$ is Full minus Text-only in pp. Coral highlights full cards; bold marks the best score per model and metric.}
\label{tab:skill-modality-ablation}
\centering
\small
\setlength{\tabcolsep}{6.0pt}
\gridsetup
\begin{tabular}{|l|r|r|r|r|r|}
\hline
\rowcolor{HeaderGray}
\textbf{VSC condition} &
\multicolumn{2}{c|}{\textbf{Mind2Web}} &
\textbf{WebLINX-BG} &
\multicolumn{2}{c|}{\textbf{OSWorld-G}} \\
\cline{2-6}
\rowcolor{HeaderGray}
& \textbf{Step SR $\uparrow$} & \textbf{Elem. Acc. $\uparrow$} &
\textbf{Overall $\uparrow$} &
\textbf{Ground. Acc. $\uparrow$} & \textbf{BBox Hit $\uparrow$} \\
\hline
\rowcolor{HeaderGray}
\multicolumn{6}{|l|}{\textbf{Qwen3-VL-2B}} \\
\hline
\textcolor{SeedBlue}{\ding{45}}~\textbf{Text-only} (w/o VSC images) & 14.0 & 67.5 & 11.0 & 25.5 & 24.0 \\
\hline
\rowcolor{CoralTint}
\evidencecards~\textcolor{CoralDark}{\textbf{Full SkillLens}} & \textbf{18.5} & \textbf{73.0} & \textbf{16.5} & 41.0 & 38.0 \\
\textcolor{MutedText}{$\Delta$ Full $-$ Text-only} & \posdelta{+4.5} & \posdelta{+5.5} & \posdelta{+5.5} & \posdelta{+15.5} & \posdelta{+14.0} \\
\hline
\evidenceboard~\textbf{Images-only} (w/o VSC text) & 12.5 & 70.0 & 15.5 & \textbf{43.0} & \textbf{39.0} \\
\specialrule{0.65pt}{0pt}{0pt}
\rowcolor{HeaderGray}
\multicolumn{6}{|l|}{\textbf{Gemini 2.5 Flash}} \\
\hline
\textcolor{SeedBlue}{\ding{45}}~\textbf{Text-only} (w/o VSC images) & 75.0 & 86.0 & 12.8 & 10.5 & 10.5 \\
\hline
\rowcolor{CoralTint}
\evidencecards~\textcolor{CoralDark}{\textbf{Full SkillLens}} & \textbf{79.0} & \textbf{89.5} & \textbf{16.0} & \textbf{27.5} & \textbf{26.0} \\
\textcolor{MutedText}{$\Delta$ Full $-$ Text-only} & \posdelta{+4.0} & \posdelta{+3.5} & \posdelta{+3.2} & \posdelta{+17.0} & \posdelta{+15.5} \\
\hline
\evidenceboard~\textbf{Images-only} (w/o VSC text) & 75.5 & 87.5 & 14.0 & 11.5 & 11.0 \\
\hline
\end{tabular}
\gridreset
\end{table*}
\endgroup

\noindent\textbf{Score differences vary by executor and task.}
Gemini 2.5 Flash's reference configuration has the largest Mind2Web Step SR difference (79.0 versus 57.9, +21.1). On WebLINX-BG, its Intent Match is 64.5 versus 49.0 and Overall is 16.0 versus 7.6, with Element IoU unchanged at 11.0. Qwen3-VL-2B's reference configuration scores +9.7 Intent Match and +3.1 IoU above No-skill. Both Gemini models produce 100\% valid OSWorld-G outputs in either condition, placing their grounding differences within valid responses.

\noindent\textbf{Paired grounding outcomes show more rescues than regressions.}
On GPT-5.4-mini's 200 OSWorld-G pairs, the reference configuration corrects 62 No-skill errors and regresses on 19 previously correct cases, yielding 43 additional correct predictions with fixed executor parameters. The 21.5-point difference has a paired percentile case-bootstrap 95\% confidence interval (CI) of [13.0, 30.0]. Appendix~\ref{app:runtime-reliability} gives the retrospective paired-case protocol and record provenance. These outcomes describe the complete reference configuration; RQ2 varies its card content.

\begin{findingbox}[Finding 1]
\textbf{Reference configurations achieve higher primary point scores across all six executors} under the settings in Section~\ref{sec:runtime-evaluation-settings}.
\end{findingbox}

\subsection{RQ2: Which VSC Content Helps?}
\label{sec:tutorial-content}
\begin{table}[!htbp]
\caption{Training-source VSC control: 64 tasks, pre-action queries, and strict operation/target/value accuracy. $\Delta$ is relative to Text-only in pp.}
\label{tab:local-tutorial-control}
\centering
\small
\setlength{\tabcolsep}{5.0pt}
\gridsetup
\begin{tabular}{|l|r|r|r|}
\hline
\rowcolor{HeaderGray}
\textbf{Memory} & \textbf{Correct} & \textbf{Acc. (\%) $\uparrow$} & \textbf{$\Delta$ (pp)} \\
\hline
\evabsent~No memory & 3/64 & 4.69 & $-1.56$ \\
\hline
\evidenceboard~Raw example & 5/64 & 7.81 & $+1.56$ \\
\hline
\textcolor{SeedBlue}{\ding{45}}~Text-only VSC & 4/64 & 6.25 & -- \\
\hline
\rowcolor{CoralTint}
\evidencecards~\textcolor{CoralDark}{\textbf{Full VSC}} & \textbf{12/64} & \textbf{18.75} & \textbf{$+12.50$} \\
\hline
\end{tabular}
\gridreset
\end{table}

\noindent\textbf{Illustrated procedures improve Web action prediction.}
Table~\ref{tab:skill-modality-ablation} compares full VSCs with text-only and images-only cards. On Mind2Web, Qwen3-VL-2B improves from 14.0 to 18.5 Step SR (+4.5), and Gemini 2.5 Flash improves from 75.0 to 79.0 (+4.0). Element Acc. also rises, from 67.5 to 73.0 for Qwen and from 86.0 to 89.5 for Gemini. Step SR jointly checks the target and operation/value.

Full cards also exceed images-only by 6.0/3.5 Step SR points for Qwen/Gemini. For Qwen, images-only improves Element Acc. over text-only (70.0 versus 67.5) while lowering Step SR (12.5 versus 14.0); combining both reaches 73.0 and 18.5. On WebLINX-BG, Full leads text-only by 5.5/3.2 Overall points and images-only by 1.0/2.0. Together, these comparisons support combining procedural text and visual references for complete Web actions.

\noindent\textbf{Desktop grounding exposes executor-specific content preferences.}
On OSWorld-G, full cards exceed text-only cards by 15.5 points for Qwen (41.0 versus 25.5) and 17.0 for Gemini (27.5 versus 10.5). Text-only scores fall below No-skill (28.5/17.5), while full-card scores exceed it. Images alone reach 43.0 for Qwen and 11.5 for Gemini. Gemini favors combined context, while Qwen favors images alone in this desktop diagnostic.

\begingroup\begin{table*}[t]
\caption{Retrieval-only coverage (0--1); parentheses give changes from Text-indexed (base). Oracle coverage refers to the source library; OSWorld-G uses same-instance case-ID recovery. Bold italics mark benchmark maxima. OCR+Text uses DeepSeek-OCR transcription and reranks up to five rows.}
\label{tab:retrieval-coverage}
\centering
\small
\begingroup
\setlength{\tabcolsep}{4.1pt}
\gridsetup
\color{black}
\setlength{\fboxsep}{1.2pt}
\setlength{\fboxrule}{0.3pt}
\renewcommand{\selectortag}[3]{\fcolorbox{BorderGray}{white}{\textcolor{#1}{#2}\,\textcolor{black}{#3}}}
\newcommand{\retrievalbest}[1]{\textbf{\textit{#1}}}
\begin{tabular}{|l|l|l|r|r|r|r|r|}
\hline
\textbf{Benchmark} & \textbf{Selector} & \textbf{Backbone} &
\textbf{Oracle Cov.\,$\uparrow$} &
\textbf{Hit@1 / $\Delta$ $\uparrow$} &
\textbf{Hit@3 / $\Delta$ $\uparrow$} &
\textbf{Hit@5 / $\Delta$ $\uparrow$} &
\textbf{Exp. Hit / $\Delta$ $\uparrow$} \\
\hline
\multirow{3}{*}{\textbf{Mind2Web}} & \textbf{\textselector} & -- & 1.00 & 0.86 (base) & 0.90 (base) & 0.91 (base) & 0.90 (base) \\
\cline{2-8}
 & \textbf{\vlmselector} & GPT-4o & 1.00 & \retrievalbest{0.93} (\posdelta{+0.07}) & \retrievalbest{0.93} (\posdelta{+0.03}) & \retrievalbest{0.93} (\posdelta{+0.02}) & \retrievalbest{0.93} (\posdelta{+0.03}) \\
\cline{2-8}
 & \textbf{\ocrselector} & DeepSeek-OCR & 1.00 & 0.87 (\posdelta{+0.01}) & 0.88 (\negdelta{-0.02}) & 0.88 (\negdelta{-0.03}) & 0.88 (\negdelta{-0.02}) \\
\hline
\multirow{3}{*}{\textbf{WebLINX-BG}} & \textbf{\textselector} & -- & 1.00 & 0.40 (base) & 0.56 (base) & 0.64 (base) & 0.50 (base) \\
\cline{2-8}
 & \textbf{\vlmselector} & GPT-4o & 1.00 & \retrievalbest{0.55} (\posdelta{+0.15}) & \retrievalbest{0.61} (\posdelta{+0.05}) & \retrievalbest{0.65} (\posdelta{+0.01}) & \retrievalbest{0.58} (\posdelta{+0.08}) \\
\cline{2-8}
 & \textbf{\ocrselector} & DeepSeek-OCR & 1.00 & 0.32 (\negdelta{-0.08}) & 0.48 (\negdelta{-0.08}) & 0.52 (\negdelta{-0.12}) & 0.52 (\posdelta{+0.02}) \\
\hline
\multirow{3}{*}{\textbf{OSWorld-G}} & \textbf{\textselector} & -- & 1.00 & \retrievalbest{1.00} (base) & \retrievalbest{1.00} (base) & \retrievalbest{1.00} (base) & \retrievalbest{1.00} (base) \\
\cline{2-8}
 & \textbf{\vlmselector} & GPT-4o & 1.00 & \retrievalbest{1.00} (\zerodelta{+0.00}) & \retrievalbest{1.00} (\zerodelta{+0.00}) & \retrievalbest{1.00} (\zerodelta{+0.00}) & \retrievalbest{1.00} (\zerodelta{+0.00}) \\
\cline{2-8}
 & \textbf{\ocrselector} & DeepSeek-OCR & 1.00 & 0.92 (\negdelta{-0.08}) & 0.97 (\negdelta{-0.03}) & 0.97 (\negdelta{-0.03}) & 0.97 (\negdelta{-0.03}) \\
\hline
\end{tabular}\gridreset
\endgroup
\end{table*}
\endgroup
\begingroup\begin{table*}[t]
\caption{GPT-5.4-mini relevance controls (200 cases per condition; \%). Deltas use Retrieved VSC (coral). Random cards are uniform samples; irrelevant cards have low query overlap. Bold marks best task/target scores.}
\label{tab:random-irrelevant-vsc-control}
\centering
\small
\setlength{\tabcolsep}{6.0pt}
\gridsetup
\begin{tabular}{|l|r|r|r|r|r|r|r|r|}
\hline
\rowcolor{HeaderGray}
\textbf{Condition} &
\multicolumn{4}{c|}{\textbf{Mind2Web: Exact Step Acc. $\uparrow$}} &
\multicolumn{4}{c|}{\textbf{OSWorld-G: Grounding Acc. $\uparrow$}} \\
\cline{2-9}
\rowcolor{HeaderGray}
\textbf{} & \textbf{Step} & \textbf{Target} & \textbf{Valid} & \textbf{$\Delta$}
& \textbf{Ground.} & \textbf{BBox} & \textbf{Valid} & \textbf{$\Delta$} \\
\hline
\evabsent~No-skill & 76.5 & 85.5 & 100.0 & \negdelta{-15.5} & 45.0 & 43.0 & 100.0 & \negdelta{-21.5} \\
\hline
\rowcolor{CoralTint}
\evidencecards~\textcolor{CoralDark}{\bfseries Retrieved VSC} & \bestscore{92.0} & \bestscore{93.0} & 100.0 & Ref. & \bestscore{66.5} & \bestscore{65.0} & 100.0 & Ref. \\
\hline
\textcolor{MutedText}{\ding{111}}~Random VSC & 72.0 & 76.5 & 100.0 & \negdelta{-20.0} & 10.5 & 10.0 & 99.0 & \negdelta{-56.0} \\
\hline
\textcolor{FakeRed}{\ding{111}}~Irrelevant VSC & 71.0 & 74.5 & 100.0 & \negdelta{-21.0} & 12.5 & 12.0 & 99.0 & \negdelta{-54.0} \\
\hline
\end{tabular}
\gridreset

\end{table*}
\endgroup
\begingroup\begin{table}[!htbp]
\caption{Selector ablation: 200 WebLINX-BG turns, frozen Qwen3-VL-2B; both selectors expand full Visual Skill Cards (VSCs). Scores (\%), latency (s).}
\label{tab:retrieval-ablation}
\centering
\small
\setlength{\tabcolsep}{4.0pt}
\gridsetup
\begin{tabular}{|l|r|r|}
\hline
\rowcolor{HeaderGray}
\textbf{Metric} &
\textbf{\textselector} &
\textbf{\vlmselector} \\
\hline
\textbf{Overall $\uparrow$} & \cellcolor{SeedTint}\bestcell{17.45} & 13.58 \\
\hline
\textbf{IM $\uparrow$} & \cellcolor{SeedTint}\bestcell{67.44} & 63.95 \\
\hline
\textbf{Ele IoU $\uparrow$} & 16.96 & \cellcolor{BoardTint}\bestcell{18.98} \\
\hline
\textbf{Dialog Acc. $\uparrow$} & \cellcolor{SeedTint}\bestcell{54.50} & 50.00 \\
\hline
\textbf{Latency $\downarrow$} & \cellcolor{SeedTint}\bestcell{1.5\,s} & 4.0\,s \\
\hline
\end{tabular}\gridreset

\end{table}
\endgroup
\begin{table}[!htbp]
\caption{Teacher-context controls (\%) with the same 512 training examples and 128 updates; students use native inputs without runtime cards. (a) 200 test-task steps: three-seed mean $\pm$ sample standard deviation; Base evaluated once. (b) 119 tasks, ten training-disjoint websites, seed 42. Full Visual Skill Card (Full-VSC) teachers are coral.}
\label{tab:carddistill-attribution-main}
\centering
\small
\setlength{\tabcolsep}{3.0pt}
\gridsetup
\begin{tabular}{|l|r|r|r|}
\hline
\rowcolor{HeaderGray}
\multicolumn{4}{|l|}{\textbf{(a) Test-task cohort: three training seeds}} \\
\hline
\textbf{Student / teacher} & \textbf{Step SR $\uparrow$} & \textbf{Ele. Acc. $\uparrow$} & \textbf{Action F1 $\uparrow$} \\
\hline
Base & 3.5 & 13.5 & 48.3 \\
\hline
\textcolor{SeedBlue}{\ding{45}} Compact text & $13.0\pm0.0$ & $19.2\pm1.3$ & $80.3\pm0.9$ \\
\hline
\rowcolor{CoralTint}
\textcolor{CoralDark}{\ding{72}\ \textbf{Full-VSC}} & $\mathbf{21.7}\pm1.6$ & $\mathbf{26.8}\pm1.8$ & $\mathbf{85.8}\pm0.7$ \\
\hline
\end{tabular}
\par\vspace{4pt}
\begin{tabular}{|l|r|r|r|}
\hline
\rowcolor{HeaderGray}
\multicolumn{4}{|l|}{\textbf{(b) Cross-website cohort: seed 42}} \\
\hline
\textbf{Student / teacher} & \textbf{Step SR $\uparrow$} & \textbf{Ele. Acc. $\uparrow$} & \textbf{Action F1 $\uparrow$} \\
\hline
Base & 3.4 & 15.1 & 47.1 \\
\hline
\textcolor{SeedBlue}{\ding{45}} Compact text & 13.4 & 21.8 & 74.4 \\
\hline
\textcolor{StrokeTeal}{\ding{45}} Richer text & 10.9 & 22.7 & 65.3 \\
\hline
\textcolor{MutedText}{\ding{111}}\ w/o rendered card & 13.4 & 21.8 & 70.4 \\
\hline
\rowcolor{CoralTint}
\textcolor{CoralDark}{\ding{72}\ \textbf{Full-VSC}} & \textbf{19.3} & \textbf{25.2} & \textbf{81.8} \\
\hline
\end{tabular}
\gridreset
\end{table}

\noindent\textbf{Training-source visual cards improve strict action accuracy.}
\label{sec:local-tutorial-control}
The 64-task control (Table~\ref{tab:local-tutorial-control}) retrieves from 7,775 training actions across 1,009 tasks using pre-action queries and observable website/domain priors. Exact screenshot duplicates are excluded. Text-only and full cards share the selected source and prompt text, and all arms retain the live screen. Full cards achieve 12/64 correct actions (18.75\%) versus 4/64 (6.25\%) for text-only cards: eight additional successes, or +12.50 points. Strict success jointly checks the target, operation, and value.

Full cards exceed raw examples (5/64) by seven correct actions (+10.94 points) and No-memory (3/64) by nine (+14.06). Raw examples and full cards share the retrieved screenshot and target crop; Full organizes this evidence with procedure and applicability fields. Appendix~\ref{app:local-tutorial-control} details this study's candidate, prompt, and strict-scoring configuration.

\begin{findingbox}[Finding 2]
\textbf{Full cards lead both single modalities in Web action prediction.} They yield eight more strict successes than text-only cards in the independent 64-task control.
\end{findingbox}

\subsection{RQ3: How Do Selection and Relevance Affect Execution?}
\label{sec:retrieval-results}

\noindent\textbf{Board selection changes ranking in the reference diagnostics.}
Table~\ref{tab:retrieval-coverage} evaluates whether selectors recover relevant cards before action prediction. Rule-based oracle coverage of the source library is 100\%. VLM-board raises Hit@1 from 86\% to 93\% on Mind2Web and from 40\% to 55\% on WebLINX-BG. After expansion, the corresponding hit rates rise from 90\% to 93\% and from 50\% to 58\%. On WebLINX-BG, board selection improves early ranking more than expanded coverage. VLM-board's Hit@5 is 65\% versus 64\% for text-indexed selection, concentrating its advantage at the first rank. The OSWorld-G diagnostic includes the evaluated case in the source bank and its identifier in the query, with relevance defined by exact case identity. Its 100\% text-indexed and VLM-board scores measure same-instance card recovery.

\noindent\textbf{Retrieval coverage and executor utility diverge.}
The execution study uses a separate cohort of 200 WebLINX-BG turns with a frozen Qwen3-VL-2B executor (Table~\ref{tab:retrieval-ablation}). Text-indexed selection reaches 17.45 Overall versus 13.58 for VLM-board, with Intent Match of 67.44 versus 63.95 and dialog accuracy of 54.50 versus 50.00. Latency is 1.5 versus 4.0\,s. VLM-board has higher Element IoU, 18.98 versus 16.96, together with lower intent and aggregate scores. Both selectors expand visual VSC evidence. The text index gains 3.87 Overall points with 2.5\,s lower reported latency; higher IoU alone does not ensure better action prediction.

\noindent\textbf{OCR+Text trails text indexing on two benchmarks.}
OCR+Text uses optical character recognition (OCR) before textual reranking. Its Hit@1 versus text indexing is 87\% versus 86\% on Mind2Web, 32\% versus 40\% on WebLINX-BG, and 92\% versus 100\% on OSWorld-G. VLM-board exceeds OCR+Text on all three benchmarks.

\noindent\textbf{Unrelated cards reduce prediction accuracy.}
Table~\ref{tab:random-irrelevant-vsc-control} fixes GPT-5.4-mini, the prompt template, and expansion budget while substituting random or low-overlap cards. On 200 Mind2Web cases, retrieved cards achieve 92.0 exact Step Acc., compared with 76.5 for No-skill, 72.0 for random cards, and 71.0 for irrelevant cards. This control uses its own subset and exact-action metric. The corresponding OSWorld-G scores are 66.5, 45.0, 10.5, and 12.5. Output validity remains 99--100\% on OSWorld-G, placing the grounding drop within mostly valid responses. These controls describe the relevance sensitivity of the reference configuration. The direction of the reference comparison reverses with card relevance.

\begin{findingbox}[Finding 3]
\textbf{Card relevance matters within the reference comparisons.} Text-indexed selection of full VSCs leads VLM-board on Overall and latency.
\end{findingbox}

\subsection{RQ4: Can Card-Conditioned Training Improve a Retrieval-Free Student?}
\label{sec:carddistill-results}
\begin{figure}[!t]
\centering
\includegraphics[width=\columnwidth]{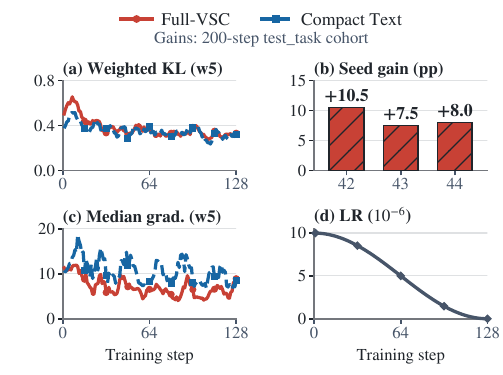}
\caption{CardDistill, seeds 42--44 (w5: trailing five-update mean). Full Visual Skill Card (Full-VSC) gains over compact text use 200 test-task steps without runtime retrieval.}
\Description{Four panels show confidence-weighted loss, per-seed accuracy gains, median gradient norm, and learning rate from the matched teacher-context training runs.}
\label{fig:carddistill-training-main}
\end{figure}

\begin{figure}[!t]
\centering
\includegraphics[width=\columnwidth,trim=0 12bp 0 14bp,clip]{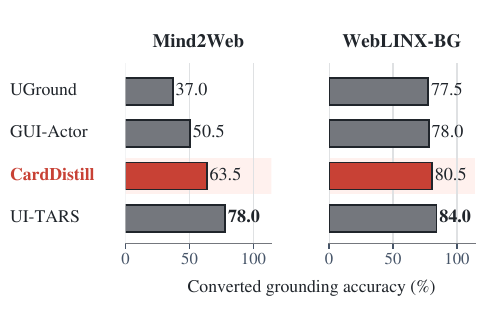}
\caption{Shared localization comparison: center-in-target-box scores on 200 examples per benchmark. Coral marks CardDistill; gray marks external grounding models.}
\Description{Horizontal bar panels compare CardDistill with external GUI grounding models on Mind2Web and WebLINX-BG under a shared center-in-target-box evaluation. Point estimates are directly labelled.}
\label{fig:grounding-baseline-profile}
\end{figure}

\begin{figure}[!t]
\centering
\includegraphics[width=\columnwidth]{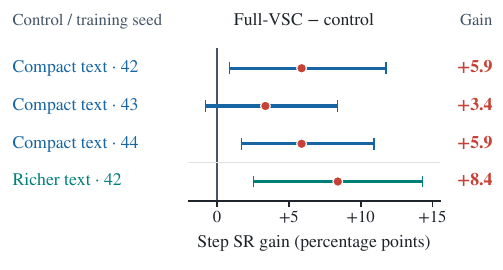}
\caption{Student step success rate gains from full Visual Skill Card (Full-VSC) versus text teachers, on 119 shared tasks from ten training-disjoint websites. Seeds are matched; bars show paired task-bootstrap 95\% intervals. Only the seed-43 interval includes zero.}
\Description{Four horizontal intervals show Full-VSC student gains over compact-text training at seeds 42, 43, and 44, and richer-text training at seed 42. All point estimates are positive; the compact-text seed-43 interval crosses zero.}
\label{fig:crosswebsite-transfer}
\end{figure}

\noindent\textbf{Matched teacher-context training improves transfer to unseen websites.}
With the same 512 training examples and 128 updates, Full-VSC teachers improve retrieval-free students over compact-text teachers on 119 tasks, one step each, from ten training-disjoint websites. The three seeds gain 5.88, 3.36, and 5.88 Step SR points, averaging 5.04; Element Acc. gains span 1.68--5.04 points and Action F1 gains span 6.18--9.74. This comparison evaluates the complete teacher-context package, combining visual focus, layout, parameter binding, and context length.

At seed 42, Full-VSC reaches 19.3 Step SR, versus 10.9 for richer text and 13.4 for either compact text or no rendered card (Table~\ref{tab:carddistill-attribution-main}b). Richer text improves Element Acc. over compact text (22.7 versus 21.8) but reduces Action F1 (65.3 versus 74.4); Full-VSC leads both.

\noindent\textbf{Full-VSC teachers also improve students on the test-task cohort.}
Under the same training controls, Full-VSC reaches $21.7\pm1.6$ Step SR versus $13.0\pm0.0$ for compact text on 200 steps from 120 tasks (Table~\ref{tab:carddistill-attribution-main}a). Seed-wise gains are 10.5/7.5/8.0 points. Element Acc. rises from 19.2 to 26.8 and Action F1 from 80.3 to 85.8. Figure~\ref{fig:carddistill-training-main} shows training traces.

\noindent\textbf{Cross-website gains persist in sensitivity checks.}
Versus richer text, 12 rescues and two regressions yield +8.4 points (paired task-bootstrap 95\% CI [2.52, 14.29]; website-cluster CI [1.82, 15.79]). Leave-one-website-out gains remain positive (6.36--10.09 points); excluding 26 pre-flagged near-similar cases yields 10.75 points on 93 cases. Figure~\ref{fig:crosswebsite-transfer} shows paired gains and uncertainty; the seed-43 compact-text interval spans zero.

\noindent\textbf{Aligned cards outperform shuffled cards.}
On the original cohorts, CardDistill raises Mind2Web Step SR from 17.5 to 29.5 (+12.0; CI [7.5, 17.5]) and WebLINX-BG Overall from 16.1 to 19.3 (+3.2; CI [0.88, 5.85]). Mind2Web plain distillation scores 17.0 and shuffled cards 19.0. CardDistill's 10.5-point gain over shuffling (CI [5.5, 15.5]) links learning to state--evidence alignment.

\noindent\textbf{Specialized models provide a grounding reference.}
CardDistill reaches 63.5/80.5 on Mind2Web/WebLINX-BG, alongside GUI-Actor's 50.5/78.0 and UI-TARS's 78.0/84.0~\citep{wu2025guiactor,qin2025uitars} (Figure~\ref{fig:grounding-baseline-profile}); these models use different training resources and input interfaces.

\begin{findingbox}[Finding 4]
\textbf{VSC teacher context benefits retrieval-free students.} On unseen websites, the complete Full-VSC package gains 5.04 points over compact text across three seeds and 8.40 over richer text at seed 42.
\end{findingbox}

\section{Conclusion}
SkillLens unifies procedural text and visual evidence in a shared representation for runtime memory and teacher supervision. With matched training data and updates, CardDistill's complete Full-VSC teacher-context package improves retrieval-free students over compact text across three seeds, with a mean Step SR gain of 5.04 points on 119 tasks from ten training-disjoint websites. In the independent pre-action runtime control, full cards achieve 12/64 strict successes versus 4/64 for text-only cards using training-source memory. Future work can test this retrieval setting on larger source-disjoint cohorts and longer interactions.\label{www:main-text-end}
\par

\FloatBarrier
\bibliographystyle{ACM-Reference-Format}
\setlength{\bibsep}{0pt}
\bibliography{references}

\ifdefined\WWWIncludeFullAppendix\clearpage\fi
\appendix
\section{CardDistill Training Configuration}

\label{app:submission-training}

This subsection records the implemented configuration for the reported CardDistill checkpoints, including the trainable parameterization, on-policy objective, student and teacher inputs, and saved training outputs.

\begin{algorithm}[H]
\caption{CardDistill training. The teacher sees Visual Skill Card (VSC) evidence during distillation, whereas the student and final evaluator do not use runtime card retrieval.}
\label{alg:carddistill}
\textbf{Input}: Training states $\{(x,o_t,h_t)\}$ with source-aligned VSC evidence, student $\pi_\theta$, and teacher $\pi_\phi$\\
\textbf{Output}: Distilled student $\pi_\theta$
\begin{algorithmic}[1]
\STATE Load the source-aligned VSC bundle $(S_t^{\mathrm{priv}},E_t^{\mathrm{priv}})$ for $\xi_t=(x,o_t,h_t)$.
\STATE Form the privileged teacher context $\tilde{\xi}_t=(\xi_t,S_t^{\mathrm{priv}},E_t^{\mathrm{priv}})$.
\STATE Sample an on-policy response $\hat{y}_t\sim\pi_\theta(\cdot\mid\xi_t)$ without VSC evidence.
\STATE Query the teacher on prefixes $\hat{y}_{t,<j}$ with context $\tilde{\xi}_t$.
\STATE Update the student with confidence-weighted $\mathrm{KL}(\pi_\theta\|\pi_\phi)$ supervision.
\STATE Discard the teacher and VSC evidence; deploy $\pi_\theta(\cdot\mid\xi_t)$.
\end{algorithmic}
\end{algorithm}

\begin{table*}[t]
\caption{Implemented CardDistill training protocol and recorded outputs. Values are taken from the saved run configurations and trainer histories. Visual Skill Card (VSC) fields are teacher-only during training; student-only evaluation loads the exported low-rank adaptation (LoRA) adapter with the benchmark evaluation prompt and no VSC fields. The OSWorld-G training corpus excludes 54 source cases without a usable bounding box.}
\label{tab:carddistill-training-config}
\centering
\small
\gridsetup
\setlength{\tabcolsep}{5pt}
\begin{tabularx}{\textwidth}{>{\raggedright\arraybackslash}p{0.16\textwidth}*{3}{>{\raggedright\arraybackslash}X}}
\toprule
\rowcolor{HeaderGray}
\textbf{Aspect} & \textbf{Multimodal-Mind2Web} & \textbf{WebLINX-BG} & \textbf{OSWorld-G} \\
\midrule
\multicolumn{4}{l}{\textcolor{CoralDark}{\textbf{Shared training configuration}}} \\
\midrule
\textbf{Backbone and adapter} & \multicolumn{3}{>{\raggedright\arraybackslash}p{\dimexpr0.84\textwidth-4\tabcolsep\relax}}{Qwen3-VL-2B-Instruct for both student and teacher; LoRA on all linear modules (rank 8, $\alpha=32$, dropout 0.05).} \\
\addlinespace[2pt]
\textbf{Objective} & \multicolumn{3}{>{\raggedright\arraybackslash}p{\dimexpr0.84\textwidth-4\tabcolsep\relax}}{On-policy reverse KL $\mathrm{KL}(P_\theta\|P_\phi)$ on student-generated prefixes; effective temperature $T=1.0$, $\beta=1$, $\lambda=1$, and no additional supervised action loss.} \\
\addlinespace[2pt]
\textbf{Optimization} & \multicolumn{3}{>{\raggedright\arraybackslash}p{\dimexpr0.84\textwidth-4\tabcolsep\relax}}{AdamW ($\beta_1=0.9$, $\beta_2=0.95$), learning rate $10^{-5}$, cosine schedule, weight decay 0.1, no warmup, seed 42, BF16, batch size 1, and gradient accumulation 1.} \\
\addlinespace[2pt]
\textbf{Runtime} & \multicolumn{3}{>{\raggedright\arraybackslash}p{\dimexpr0.84\textwidth-4\tabcolsep\relax}}{Maximum sequence length 8192; maximum completion length 128; one training process; non-vLLM rollout path; final LoRA checkpoint retained.} \\
\midrule
\multicolumn{4}{l}{\textcolor{CoralDark}{\textbf{Dataset-specific training inputs and outputs}}} \\
\midrule
\textbf{Training rows / steps / final loss} & 512 / 128 / 0.1829 & 512 / 128 / 0.1216 & 510 / 141 / 0.1328 \\
\addlinespace[2pt]
\textbf{Student prompt} & Task, site/domain, up to four preceding actions, up to 16 candidates, 1,800-character HTML, and live screenshot. & Demo/turn metadata, viewport, dialogue, action history, up to 20 candidates, 1,800-character clean HTML, and live screenshot. & Instruction and live screenshot. \\
\addlinespace[2pt]
\textbf{Student output} & JSON \texttt{\{op, backend\_node\_id, value\}}. & Benchmark-native structured browser action. & Normalized \texttt{computer\_use} tool call. \\
\addlinespace[2pt]
\textbf{Teacher-only VSC bundle} & Compact VSC text (name, goal, first three procedure steps, up to 12 keywords) and 4--5 image paths in total, including the live screen. & Compact VSC text and 3 image paths in total, including the live screen. & Compact VSC text and 3 image paths in total, including the live screen. \\
\addlinespace[2pt]
\textbf{Saved output} & LoRA checkpoint-128 & LoRA checkpoint-128 & LoRA checkpoint-141 \\
\bottomrule
\end{tabularx}
\gridreset

\end{table*}

\section{Metric Definitions}
\label{app:submission-metrics}

For Mind2Web, let $e_i=\mathbf{1}[\hat e_i\in\mathcal P_i]$ indicate whether the predicted target is positive. Let $A_i$ and $\widehat A_i$ be whitespace-token sets of the reference and predicted operation/value strings. The metrics are
\begin{subequations}
\begin{align}
F_{1,i} &= \frac{2|A_i\cap\widehat A_i|}{|A_i|+|\widehat A_i|}, \\
\mathrm{EleAcc} &= \frac1N\sum_i e_i, \\
\mathrm{ActionF1} &= \frac1N\sum_i F_{1,i}, \\
\mathrm{StepSR} &= \frac1N\sum_i e_i\,\mathbf{1}[F_{1,i}=1].
\end{align}
\end{subequations}
The evaluator sets $F_{1,i}=1$ when both token sets are empty; a single empty set gives zero. The operation/value serializer and token matching are unchanged in all controlled studies.

For WebLINX-BG, Intent Match is the mean indicator of correct intent. Element IoU averages $|\hat b_i\cap b_i|/|\hat b_i\cup b_i|$ over element-grounded reference actions, with $\hat b_i$ and $b_i$ the predicted and reference boxes. On text-bearing actions, the field named Text F1 averages CHRF for matching \texttt{say}/\texttt{textinput} intents and URLF for matching \texttt{load} intents; an intent mismatch scores zero. Overall averages the official per-turn score: it combines IoU and CHRF when both apply, uses URLF for load actions, uses the available IoU or CHRF otherwise, and is zero for incorrect intent. These fields retain their benchmark-specific definitions.

For OSWorld-G, a prediction is correct when its point lies inside the annotated box/polygon, equals an annotated point target, or predicts refusal for a refusal case. Grounding Acc. is the mean of these per-case correctness indicators. It measures localization, not multi-step desktop-task completion.

\section{Limitations}

\label{app:submission-limitations}

Runtime results depend on card provenance and the information supplied to selection. Historical mappings between cards and tests are incomplete for case and task IDs, screenshot and DOM hashes, and page-scoped targets. Source-library coverage and shared benchmark names do not establish source-disjoint transfer. The independent 64-task study evaluates pre-action retrieval from training-source memory, with a 12.50-point Full-versus-Text accuracy difference. Its sample size and distinct candidate, prompt, and scoring configuration delimit comparisons with the broader reference protocols. CardDistill depends on training-card quality and the student policy. Action serialization and live-screen grounding remain responsibilities of the executor. Specialized grounding mechanisms are complementary to SkillLens, motivating separate coverage, action, and localization metrics.

Board diagnostics examine direct VLM selection and OCR transcription followed by textual reranking. Similar rows and ambiguous transcriptions remain difficult; the main configuration uses text-indexed selection with visual evidence expansion.

CardDistill evaluates students without retrieval on Mind2Web and WebLINX-BG; matched controls use Mind2Web. OSWorld-G evaluates runtime memory. Converted point scores remain separate web-cohort diagnostics. These offline studies do not establish multi-step completion or end-to-end cost savings. Teacher-context controls match data and updates, while visual focusing, layout, and context length remain coupled.

\noindent\textbf{Reference-experiment provenance.}
The runtime reference configurations are documented from archived adapters and reported result aggregates. Historical run payloads and complete card-to-test mappings are unavailable, leaving the exact input path for individual historical rows and source--test separation unresolved. Section~\ref{sec:runtime-evaluation-settings} defines the adapter inputs; the independent runtime control and controlled student studies have their own protocols below.

\section{Runtime Grounding Reliability}
\label{app:runtime-reliability}

We analyze the exported paired grounding counts for GPT-5.4-mini on OSWorld-G retrospectively, without new model inference. Among 200 pairs, both conditions are correct on 71, VSCs rescue 62, regress on 19, and both fail on 48. These cells recover the main-table scores: No-skill is 90/200 (45.0\%) and VSC is 133/200 (66.5\%).

For the accuracy difference, each pair contributes $d\in\{-1,0,1\}$ with empirical probabilities $(19,119,62)/200$. We compute the complete distribution of the mean of 200 bootstrap draws by convolution and take its 2.5th and 97.5th percentiles: [13.0, 30.0] percentage points. This removes Monte Carlo error, not the approximation in bootstrap confidence coverage. The two-sided exact McNemar test on the 81 discordant pairs gives $p=1.77\times10^{-6}$. As a conservative sensitivity check, two 97.5\% Clopper--Pearson intervals for rescue and regression probabilities, combined by Bonferroni and differencing endpoints, give a 95\% interval of [8.65, 33.49] points.

The calculation assumes independent cases. Raw records and cluster identifiers are unavailable for this reanalysis, so original pairing, exclusions, and app/screenshot dependence cannot be rechecked. These intervals neither reproduce the original model runs nor measure seed variability; VSC versus No-skill tests the whole memory intervention. We exclude the WebLINX-BG export because its dialog-turn accuracy differs from official Overall.

Original target-size bins show descriptive gains of +27.69 points (16 to 34 correct out of 65 small targets), +32.81 (25 to 46/64 medium), +7.94 (45 to 50/63 large), and $-12.50$ (4 to 3/8 unknown). Within-bin paired cells are unavailable; these observations do not establish subgroup differences or size-dependent effects.

\section{CardDistill Reliability and Isolation}
\label{app:carddistill-reliability}

For the separately reported original cohorts in Section~\ref{sec:carddistill-results}, intervals use 10,000 paired resamples of evaluation cases. The Mind2Web comparison uses 200 \texttt{test\_task} steps; WebLINX-BG retains the 172 turns with valid official reference scores from the 200-turn \texttt{test\_iid} cohort. The two-sided bootstrap $p$-values for CardDistill versus Base are $<.001$ and .005, respectively. On Mind2Web, Plain OPD versus Base has $p=.796$, Shuffled VSC versus Base has $p=.543$, and CardDistill versus Shuffled VSC has $p<.001$. Their Step SR gains versus Base are $-0.5$ (CI [-2.0, 1.0]), $+1.5$ ([-2.5, 5.5]), and $+12.0$ ([7.5, 17.5]) for Plain OPD, Shuffled VSC, and CardDistill, respectively. The new controlled studies use the task and website sampling protocols in Appendix~\ref{app:teacher-modality-control}.

For those original cohorts, the reported exact-identifier and SHA-256 audits compared the 512-example training sources against each benchmark's 200-case evaluation cohort before training. Both audits found zero overlap in case identifiers, task identifiers, screenshot bytes, normalized DOM content, and target identifiers. This audit addresses exact source overlap; the paired bootstrap addresses evaluation-sample uncertainty for the reported checkpoints, not variation across training seeds.

\section{Matched Teacher-Context Protocols}
\label{app:teacher-modality-control}

\textbf{Training and evidence.} The controlled studies use 512 distinct Mind2Web training tasks and eight separate development tasks. Full-VSC's reference screenshot is a copy of the current training screenshot; its target crops and rendered card contain label-derived privileged cues. The training card encodes the current labeled example. Compact text includes the operation and target IDs. The card adds explicit parameter bindings and target summaries; nine of 88 nonempty action-value strings are absent verbatim from the compact/student text. These counts describe verbatim textual presence, not visibility in the live screenshot.

Richer text appends every renderer text line and image caption in original order, omitting frontmatter and extra image pixels. It preserves Unicode and fields without truncation; the longest teacher prompt plus 128 completion tokens is 5,846, below the 8,192-token limit. Raster clipping, glyph support, and image resizing prevent strict text--pixel equivalence. All trained arms reported here share the Qwen3-VL-2B foundation checkpoint, LoRA settings, optimizer settings, 512 examples, four H800 GPUs, and 128 updates; the vision encoder and aligner are frozen. Logs verify one successful use per training example and matching per-rank orders within each seed. This matches data and updates, not FLOPs or necessarily historical LoRA initialization bytes. Each image is capped at 1,024 tokens.

\textbf{Evaluation cohorts.} The 200-step \texttt{test\_task} cohort covers 120 tasks, excludes 78 previously used local test tasks, and has no exact training/development task, screenshot, or DOM overlap. All 30 top-16 candidate misses remain. Full-VSC Step SR for seeds 42/43/44 is 23.5/20.5/21.0, versus 13.0 for each compact-text run. Figure~\ref{fig:carddistill-training-main} uses these runs: per-update mean loss and median gradient across seeds, followed by a five-update trailing mean; the shared recorded learning rate is unsmoothed.

The additional \texttt{test\_website} cohort contains 119 tasks, one eligible step each, on ten sites absent from the 73 training sites. A frozen hash ordering selects steps before inference; six steps lacking positive annotations are skipped without dropping any task. The split has 97 CLICK, 15 TYPE, and seven SELECT actions. Official DeBERTa rankings supply 16 candidates; all 30 misses remain. Exact screenshot/DOM matches to auditable local studies are excluded. Older legacy per-case logs are unavailable, so disjointness from all historical experiments is not claimed. Twenty-six near-similarity flags are frozen before inference and retained in the primary set; the 93-case exclusion is a sensitivity analysis.

\textbf{Scoring and uncertainty.} All students use identical live inputs, greedy decoding, seed 42, and at most 128 tokens. Student prompts contain no current-action answer hint; 833 hidden-label perturbations leave their text unchanged. The existing tolerant parser and literal scorer are fixed; empty CLICK values are a shared convention established in development. On the website cohort, Full-VSC scores 23/19/25 correct for seeds 42/43/44, versus compact text's 16/15/18. The paired rescues/regressions are 9/2, 6/2, and 8/1. Richer text, no-rendered-card, and Base score 13/16/4, respectively.

Paired task and website-cluster bootstraps use 10,000 resamples (seed 20261010) and recompute micro-average differences with sampled denominators. The website bootstrap uses ten clusters. Intervals are descriptive and marginal, not simultaneous family coverage; training-seed variability is reported separately. The models share the same evaluation cases. All conditions and checkpoints are fixed before the new held-out inference, and all prediction outputs are retained in the experiment archive.

\section{Matched VSC Control Protocol}
\label{app:local-tutorial-control}

The study evaluates one action per task on 64 public Multimodal-Mind2Web tasks using a training-source memory bank. All memory arms retrieve the same source from 7,775 training actions across 1,009 tasks using BM25 and observable website/domain priors. Queries use pre-action task and interface information. Current test labels are used for scoring; exact screenshot duplicates are excluded from memory selection. Released DeBERTa scores rank the live-page element candidates. The denominator includes all 64 tasks.

The four conditions vary how the retrieved experience is presented. No-memory provides the live task and screen. Raw-example adds the source interaction and its visual references. Text-only VSC presents the source as a structured procedure with applicability fields. Full VSC combines the same card text with historical visual evidence. Text-only and full VSCs share prompt text, while raw-example and full-VSC conditions share the historical screenshot and target crop. No-memory, Text-only, and Full use the same 64 tasks, frozen executor version, base task-and-screen prompt template, and scoring implementation. Every condition retains the current screenshot.

Strict success requires the correct operation, a listed positive target, and an exact normalized-value match. Correct counts for No-memory, Raw-example, Text-only VSC, and Full VSC are 3, 5, 4, and 12 out of 64, respectively (Table~\ref{tab:local-tutorial-control}). Their accuracies are 4.69\%, 7.81\%, 6.25\%, and 18.75\%. Full VSC yields eight additional successes over Text-only (+12.50 points), seven over Raw-example (+10.94), and nine over No-memory (+14.06). These comparisons evaluate the complete action, including operation and value, alongside target selection. They support combining the structured procedure and visual reference in this training-source setting.

\section{Retrieval and Expansion Details}
\label{app:retrieval-details}
For execution cards without dependencies, the shared index uses descriptor $d_i$: identifier, name, level, goal, keywords, modal context, and dependency identifiers. Let $n(u;r)$ count token $u$ in $r$ and $\operatorname{supp}$ denote unique-token support. The scores are
\begin{equation}
\begin{aligned}
\mathcal U_{t,i}&=\operatorname{supp}(\mathrm{Tok}(q_t))\cap\operatorname{supp}(\mathrm{Tok}(d_i)),\\
r_i^{(0)}&=\sum_{u\in\mathcal U_{t,i}}\min(n(u;q_t),n(u;d_i)),\\
\mathcal M_{t,i}&=\operatorname{supp}(\mathrm{Tok}(q_t))\cap\operatorname{supp}(\mathrm{Tok}(d_i^{\mathrm{cand}})),\\
r_i^{(1)}&=|\mathcal M_{t,i}|+0.1r_i^{(0)}.
\end{aligned}
\end{equation}
Candidate descriptor $d_i^{\mathrm{cand}}$ stores identifier, name, goal, keywords, and visual tags. The Mind2Web reference adapter orders cards by site tier, operation compatibility, lexical score, and identifier. Its input conditions follow Section~\ref{sec:runtime-evaluation-settings}; Appendix~\ref{app:local-tutorial-control} details the independent control.

Equation~\ref{eq:selection-stages} specifies the web-path budgets and post-cap gate. OSWorld-G ranks in one stage without $K_s$. Card and image budgets remain fixed within each comparison.

Beyond text indexing, the default path loads selected evidence at cost $C(E_t)$; board diagnostics also construct and inspect $\mathcal B(\mathcal C_t)$. OCR+Text uses DeepSeek-OCR and reranks up to five row transcriptions, separately from direct OCR-box localization. All selectors share the identifier-to-asset expansion map.

The archived OSWorld-G diagnostic uses overlapping prefixes of source and evaluation records, a query containing the case ID, and an exact case-ID oracle. Its scores measure same-instance identity recovery; source exclusion and removal of identity hints require a new evaluation.

Web oracles define relevance by site and operation or intent agreement, using the inputs in Section~\ref{sec:runtime-evaluation-settings}.

\end{document}